\documentclass[lettersize,journal]{IEEEtran}
\usepackage{amsmath,amsfonts}
\usepackage{algorithm}
\usepackage{array}
\usepackage[caption=false,font=normalsize,labelfont=sf,textfont=sf]{subfig}
\usepackage{textcomp}
\usepackage{stfloats}
\usepackage{url}
\usepackage{verbatim}
\usepackage{graphicx}
\usepackage{cite}

\usepackage{bm} 
\usepackage{mathtools} 
\usepackage{amssymb}
\usepackage[noend]{algpseudocode} 
\algnewcommand\algorithmicforeach{\textbf{for each}}
\algdef{S}[FOR]{ForEach}[1]{\algorithmicforeach\ #1\ \algorithmicdo}
\usepackage{multirow}
\usepackage{tabularx}
\usepackage{booktabs}
\usepackage{pifont}
\usepackage{xcolor} 
\usepackage{marvosym}
\definecolor{green}{rgb}{0.0, 0.5, 0.0}

\begin{document}

\title{Rethinking Self-training for Semi-supervised Landmark Detection: A Selection-free Approach}

\author{Haibo Jin, Haoxuan Che, and Hao Chen
\thanks{This work was supported by the Hong Kong Innovation and Technology Fund (Project No. MHP/002/22), Shenzhen Science and Technology Innovation Committee Fund (Project No. SGDX20210823103201011), Research Grants Council of the Hong Kong (Project No. T45-401/22-N) and the Project of Hetao Shenzhen-Hong Kong Science and Technology Innovation Cooperation Zone (HZQB-KCZYB-2020083). \textit{(Corresponding author: Hao Chen.)}} 
\thanks{H. Jin and H. Che are with Department of Computer Science and Engineering, Hong Kong University of Science and Technology, Hong Kong, China (e-mail: \{hjinag,hche\}@cse.ust.hk)}
\thanks{H. Chen is with the Department of Computer Science and Engineering, Department of Chemical and Biological Engineering and Division of Life Science, Hong Kong University of Science and Technology, Hong Kong, China; HKUST Shenzhen-Hong Kong Collaborative Innovation Research Institute, Futian, Shenzhen, China (e-mail: jhc@cse.ust.hk)}
\thanks{H. Jin and H. Che contributed equally to this work.} 
}



\maketitle

\begin{abstract}
Self-training is a simple yet effective method for semi-supervised learning, during which pseudo-label selection plays an important role for handling confirmation bias. Despite its popularity, applying self-training to landmark detection faces three problems: 1) The selected confident pseudo-labels often contain data bias, which may hurt model performance; 2) It is not easy to decide a proper threshold for sample selection as the localization task can be sensitive to noisy pseudo-labels; 3) coordinate regression does not output confidence, making selection-based self-training infeasible. To address the above issues, we propose Self-Training for Landmark Detection (STLD), a method that does not require explicit pseudo-label selection. Instead, STLD constructs a task curriculum to deal with confirmation bias, which progressively transitions from more confident to less confident tasks over the rounds of self-training. Pseudo pretraining and shrink regression are two essential components for such a curriculum, where the former is the first task of the curriculum for providing a better model initialization and the latter is further added in the later rounds to directly leverage the pseudo-labels in a coarse-to-fine manner. Experiments on three facial and one medical landmark detection benchmark show that STLD outperforms the existing methods consistently in both semi- and omni-supervised settings. The code is available at https://github.com/jhb86253817/STLD. 
\end{abstract}


\IEEEpeerreviewmaketitle

\section{Introduction}
\label{sec1}

\IEEEPARstart{L}{andmark} detection is a vision task that aims to localize predefined points on a given image. Its application covers several areas such as facial~\cite{WQY18,FKA18,WLZ23} and medical landmark detection~\cite{ZLZ19,ZYX21,JCC23}, both of which can be the preceding steps to other tasks (e.g., face recognition~\cite{LJL13}, face editing~\cite{LMY23}, surgery planning~\cite{CMC19,LPC23}, etc.). Thus, it is essential for a landmark model to accurately and robustly localize points in the wild.

\begin{figure*}[t]
\centering
  \includegraphics[width=0.95\linewidth]{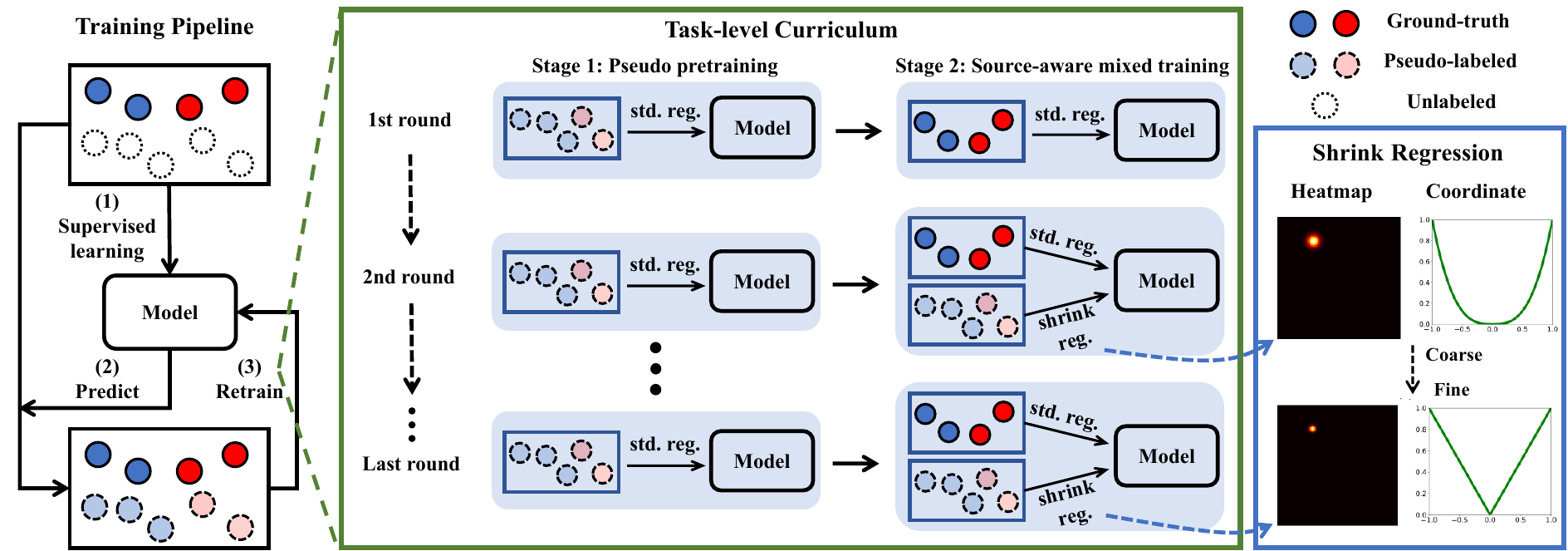}
\caption{Self-Training for Landmark Detection (STLD). (1) Model is first trained on the labeled data with supervised learning, (2) then estimates pseudo-labels of unlabeled data, and (3) is retrained on both labeled and pseudo-labeled data with the constructed task curriculum.} 
\label{fig:stld}      
\end{figure*}

Despite the progress of landmark detection, most methods~\cite{WBF19,LLZ20,KMM20} rely on fully labeled data to conduct supervised learning, which is not scalable under the big data scenario. Different from image-level labels, the annotation of landmarks requires pixel-level location of tens of points for each image, which costs much more labor. In particular, such labels of medical images are even more expensive to obtain as medical expertise is required. To address the above issue, semi-supervised learning (SSL) is often adopted for exploiting both labeled and unlabeled data. Among the existing SSL methods, self-training~\cite{Scu65,Yar95,Lee13} and its variants~\cite{XLH20,SZL20,YZQ22} are arguably the most widely used due to their simplicity and effectiveness, which iteratively enlarge the training set by adding confident pseudo-labels via sample-selection strategies.   

However, when applied to landmark detection, selection-based self-training causes several problems. \textbf{First of all}, the selected confident pseudo-labels may introduce data bias due to the biased selection, which can lead to performance degradation (see Sec.~\ref{sec3.2} for more details). \textbf{Additionally}, it is not easy to decide a proper threshold for sample selection~\cite{BTQ21,WJG22}, especially for the fine-grained localization task that is more sensitive to noisy labels~\cite{DoY19}. \textbf{Furthermore}, coordinate regression-based landmark detectors~\cite{XQH22,LGR22,LJL22} have recently obtained state-of-the-art (SOTA) results by adopting transformers~\cite{VSP17,CMS20}, yet they do not output confidence, which makes selection-based self-training methods infeasible.   

In this work, we aim to rethink self-training for landmark detection by asking the following question: \textit{Is pseudo-label selection a necessary component of self-training?} We argue that a successful self-training strategy should take care of confirmation bias by utilizing reliable information first, while \textit{obtaining such information is not necessarily restricted to the selection of confident pseudo-labels}. For example, one may conduct a simpler but more confident task on pseudo-labels to extract reliable information.
 
Based on the above observation, we propose Self-Training for Landmark Detection (STLD), a method that does not rely on explicit pseudo-label selection. Instead, it leverages pseudo-labels by constructing a task curriculum, where confident tasks are conducted first for obtaining reliable information from the pseudo-labels. Specifically, the curriculum starts with \textbf{pseudo pretraining} by only leveraging pseudo-labels for providing a better initialization; then \textbf{shrink regression} is further added to the curriculum in the later rounds to directly leverage the pseudo-labels via a coarse-to-fine process. The process starts with coarse-grained regression, then progressively increases the granularity over the rounds until the same level as standard regression. In each round, STLD follows a two-stage pipeline. The first stage conducts pseudo pretraining on all the pseudo-labeled data; the second stage, termed source-aware mixed training, leverages ground-truths (GTs) and pseudo-labeled data via standard and shrink regression, respectively. The proposed method is model-agnostic and \textbf{works with both heatmap and coordinate regression}. Fig.~\ref{fig:stld} shows the training pipeline of STLD. Our experiments on three facial and one medical landmark detection benchmark demonstrate the superiority of the proposed method, showing that STLD surpasses the existing SSL methods in both semi- and omni-supervised settings. For example, in the semi-supervised setting, our method improves the performance over SOTA methods by 5.7\% on WFLW; in the omni setting, STLD outperforms its counterparts with $10\times$ less unlabeled images (200K $\rightarrow$ 20K) on 300W. We summarize our contributions as follows.
\begin{itemize}
\item We propose to rethink self-training for semi-supervised landmark detection from a different perspective by changing from the selection-based data-level curriculum to selection-free task-level curriculum. The proposed task-level curriculum has the potential to address the problems of previous methods, namely 1) the data bias issue due to sample selection, 2) the sensitivity to threshold, and 3) the infeasibility to coordinate regression-based methods.
\item Following the philosophy of task-level curriculum, we design a novel framework for semi-supervised landmark detection. The framework gradually increases the granularity of the task over self-training rounds, enabling the model to leverage the knowledge of pseudo-labels while introducing minimum noise. Pseudo pretraining is designed as the task with the lowest level of granularity by pretraining the model on all the pseudo-labeled data; shrink regression is further added in the later rounds to directly leverage the pseudo-labels by conducting the regression task in a coarse-to-fine manner. The proposed framework is model-agnostic and works with both heatmap and coordinate regression.
\item Extensive experiments are conducted to demonstrate the effectiveness of the proposed framework in both semi- and omni-supervised settings, where the datasets cover one medical and three facial landmark detection benchmarks.
\end{itemize} 

\section{Related Works}
\label{sec2}

\subsection{Self-training}
Self-training~\cite{Scu65,Yar95} is a classic SSL method, which iteratively enlarges the GT training set by adding confident pseudo-labeled samples for obtaining better performance. Due to the simplicity and effectiveness, it has been successfully applied to various vision tasks such as image recognition~\cite{BCG19,SBL20,XLH20,ZGL20,LWL21,ZWH21}, object detection~\cite{RDG18,SZL20,ZYW21,YWW21,LMH21,LWS22}, semantic segmentation~\cite{ZYK18,YLS21,YZQ22}, and so on. However, past works have mostly adopted self-training with the sample selection strategy via either fixed~\cite{SBL20,XLH20} or dynamic thresholds~\cite{BTQ21,YZQ22} to address the confirmation bias issue. In contrast, we attempt to explore another line of research by proposing a self-training method without explicit pseudo-label selection.

\subsection{Supervised Landmark Detection}
Supervised landmark detection methods can be categorized as either coordinate regression or heatmap regression. Coordinate regression~\cite{WQY18,ZSZ19,VBV18,LZH19,LJL22} directly regresses the coordinates of landmarks, while heatmap regression~\cite{LZH19,CSJ19,ZZY19,WBF19,CBG20} uses heatmaps as a proxy of coordinates for landmark prediction. Due to the proposal of various backbones for high-resolution representation~\cite{NYD16,RFB15,WSC19}, heatmap regression has been in the leading position in model performance for a long time. Later, Transformers are introduced to this task for end-to-end coordinate regression~\cite{XQH22,LGR22,LJL22}, which show superior performance to that of heatmap regression. However, these methods rely on fully labeled data for supervised learning, which is not scalable in practice. In this work, we mainly focus on semi-supervised landmark detection rather than supervised learning, where the former is much more label-efficient than the latter. It is worth noting that the proposed SSL method is applicable to both heatmap and coordinate regression. Additionally, the proposed shrink regression also involves loss function designing. The difference between shrink regression and the previous works is mainly two-fold. 1) Previous works design loss functions under supervised learning by considering label scales~\cite{RLZ19,LWH21} or learning difficulties~\cite{FKA18,WBF19} adaptively for accurate prediction; in contrast, shrink regression is designed under semi-supervised learning by constructing a task curriculum that gradually increases learning difficulties for data-efficient learning. 2) Prior works often design loss functions for either heatmap~\cite{RLZ19,WBF19,LWH21} or coordinate regression~\cite{FKA18} specifically while shrink regression is a generic method that applies to both.

\subsection{Semi-supervised Landmark Detection}
Semi-supervised landmark detection can be divided into three types. 1) Consistency regularization based methods force models to learn consistency across different geometric~\cite{HMT18,XWZ21} or style~\cite{QSW19} transformations to obtain supervision for unlabeled images. 2) Some methods attempt to learn more generalizable representations~\cite{BjC20} or reasonable predictions~\cite{RLZ19} with unlabeled data through adversarial learning. 3) Self-training based methods iteratively estimate and utilize pseudo-labels of unlabeled data. To address confirmation bias in self-training~\cite{AOA20}, Dong et al.~\cite{DoY19} designed a teacher network to assess the quality of pseudo-labels for sample selection, but it only works with small input size (64×64) due to training difficulty.  Jin et al.~\cite{JLS21} proposed a self-training method to gradually increase the difficulty of a simpler proxy task, which does not require pseudo-label selection. However, their method~\cite{JLS21} is closely tied to a specific landmark detector, thus it is inapplicable to other SOTA models. HybridMatch~\cite{KLK23} proposed to combine 1D and 2D heatmap regression to alleviate the task-oriented and training-oriented issues, respectively, based on FixMatch~\cite{SBL20}. Additionally, a confidence-based soft filtering strategy is proposed in HybridMatch to avoid thresholding by regularizing the pseudo-labeled loss with prediction confidence, which makes it also a selection-free method. However, due to the dependence on prediction confidence, HybridMatch would inevitably introduce noise as modern neural networks are rarely calibrated and tend to give overconfident predictions~\cite{GPS17,MDR21}. Compared to these existing works~\cite{DoY19,JLS21,KLK23}, our method is a more general solution that 1) applies to both heatmap and coordinate regression, and 2) relies on neither sample selection nor confidence regularization. We attribute these advantages to the carefully designed task-level curriculum.

\section{Preliminaries}
\label{sec3}

\subsection{Problem Formulation}
\label{sec3.1}

Let $\mathcal{X}_l=\{\bm{x}_1,\bm{x}_2,...,\bm{x}_{n_l}\}$ and $\mathcal{Y}_l=\{\bm{y}_1,\bm{y}_2,...,\bm{y}_{n_l}\}$ be the images and GTs of the labeled data in the training set, respectively, where $\bm{x}_i \in \mathbb{R}^{3 \times H_0 \times W_0}$ is the $i$-th image, $\bm{y}_i \in \mathbb{R}^{N \times 2}$ is the landmark coordinates of the $i$-th image, $N$ is the number of landmarks, and $n_l$ is the number of labeled data. In supervised landmark detection, we aim to learn a function $f_{\bm{\theta}}$ by minimizing the following risk:
\begin{equation}
\mathcal{R}^{\text{SL}}(f_{\bm{\theta}}) = \sum_{\mathclap{ (\bm{x}_i,\bm{y}_i) \in (\mathcal{X}_l, \mathcal{Y}_l) }} \mathcal{L} (f_{\bm{\theta}}(\bm{x}_i), {\bm{y}_i}),
\label{eqn1}
\end{equation}
where $f_{\bm{\theta}}$ is parameterized by $\bm{\theta}$ and $\mathcal{L}$ is the loss function. 

In semi-supervised setting (including the omni-setting~\cite{RDG18}), additional $n_u$ (often $n_u \gg n_l$) unlabeled images $\mathcal{X}_u=\{\bm{x}_{n_l+1},\bm{x}_{n_l+2},...,\bm{x}_{n_l+n_u}\}$ can be used to improve the generalization of $f_{\bm{\theta}}$ with the following objective function:
\begin{equation}
\mathcal{R}^{\text{SSL}}(f_{\bm{\theta}}) = \mathcal{R}^{\text{SL}}(f_{\bm{\theta}}) + \lambda \cdot \sum_{\mathclap{ \bm{x}_i \in \mathcal{X}_u }} \mathcal{G}(f_{\bm{\theta}}, \bm{x}_i),
\label{eqn2}
\end{equation}
where $\mathcal{G}$ is the loss on unlabeled images $\mathcal{X}_u$ and $\lambda$ is a balancing coefficient. There are several options for $\mathcal{G}$. For example, consistency regularization~\cite{LaA17,SJT16,TaV17} is a common technique for designing such a loss, which imposes output consistency on the same images with different augmentations. Entropy minimization is another technique that explicitly minimizes the entropy of model predictions on unlabeled images~\cite{GrB05,MMI18}. Self-training~\cite{CSZ09}, the focus of this paper, can be seen as an implicit entropy minimization method through pseudo-labeling~\cite{Lee13}, which will be detailed below.

\begin{figure}[t]
\centering
  \includegraphics[width=1\linewidth]{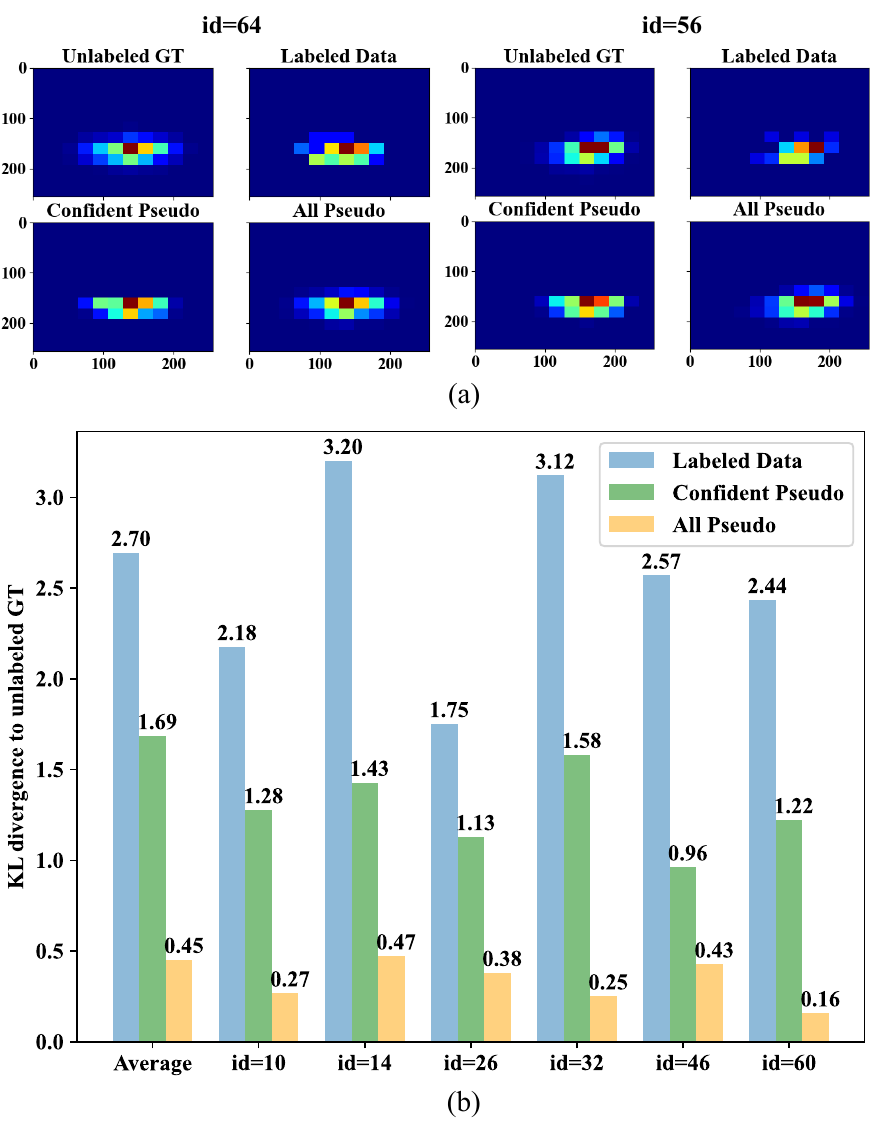}
\caption{(a) Visualized label density maps of four data groups from 300W~\cite{STZ13}. The labels (i.e., coordinates) are mapped to the $256\times256$ map, and plotted in density maps with 12 bins at each axis. (b) The KL divergence of unlabeled GT and three data groups respectively: 1) labeled data, 2) confident pseudo-labeles, and 3) all the pseudo-labels, calculated based on the label density maps over 300W~\cite{STZ13}. Both the average distance and individual distance of randomly selected landmarks are plotted. } 
\label{fig:kl}      
\end{figure}

\subsection{Selection-based Self-training}
\label{sec3.2}

To solve a semi-supervised learning problem, a typical self-training method directly solves a supervised learning problem by iteratively estimating and selecting confident pseudo-labeled samples to expand the labeled set. Its training pipeline can be described as follows: 1) Train a new model on the labeled set $(\mathcal{X}_l, \mathcal{Y}_l) \cup \mathcal{A}$ ($\mathcal{A}=\varnothing$ initially) with supervised learning; 2) use the trained model to estimate pseudo-labels $\widehat{\mathcal{Y}}_u$ for the images from unlabeled set $\mathcal{X}_u$; 3) select confident pseudo-labeled samples based on the confidence threshold $\tau$, then add the selected samples to set $\mathcal{A}$; 4) repeat steps 1 to 3 until no more data can be added. The objective function at step 1 can be written as
\begin{equation}
\mathcal{R}^{\text{SBST}}(f_{\bm{\theta}}) = \mathcal{R}^{\text{SL}}(f_{\bm{\theta}}) + \sum_{\mathclap{ (\bm{x}_i,\bm{y}_i) \in \mathcal{A} \subseteq (\mathcal{X}_u, \widehat{\mathcal{Y}}_u) }} \mathcal{L} (f_{\bm{\theta}}(\bm{x}_i), {\bm{y}_i}),
\label{eqn3}
\end{equation}
where $\mathcal{A}$ is a set containing the selected pseudo-labeled samples. As we can see from the pipeline, self-training may reinforce the model's mistakes as it attempts to teach itself (i.e., confirmation bias~\cite{AOA20}). Step 3 plays an important role in easing this issue by only retaining confident pseudo-labels, which are less likely to contain mistakes. 
 
Although selecting confident pseudo-labels mitigates the confirmation bias, it may introduce data bias. To verify this, we analyze the label distribution of different data groups via label density map. The analysis is based on 300W~\cite{STZ13}. There are four groups of labels: \textbf{1)} the GT of unlabeled data (98.4\%); \textbf{2)} the labeled data (1.6\%); \textbf{3)} all the unlabeled data with pseudo-labels estimated by the model trained on the labeled data; \textbf{4)} unlabeled data with confident pseudo-labels ($\tau=0.4$). To obtain label density maps, we first divide the 256$\times$256 map into 12$\times$12 bins, where the downsampling from 256 to 12 is to avoid the impact of small changes so that it is shift invariant to a certain degree. Then the label density maps of different groups can be obtained by calculating the frequency of landmarks within each bin, followed by normalization. Fig.~\ref{fig:kl}a visualizes the label density maps of two landmarks with ID 64 and 56. It could be visually perceived that the density map of all pseudo is more similar to that of unlabeled GT than the other two groups. To quantify the similarity, we use KL divergence to measure the distance between the anchor distribution (i.e., unlabeled GT) and the distribution of the remaining groups, respectively. KL divergence is used because it is a standard measure of the dissimilarity between two probability distributions~\cite{KuL51}.

Fig.~\ref{fig:kl}b shows the KL divergence between the unlabeled GT and the other three groups averaged over all the landmarks as well as the ones of several randomly selected landmarks. It can be seen from the figure that the labeled data has the largest KL divergence to the unlabeled GT on average, indicating that limited number of labeled data can be quite biased and thus limits the generalization of the model. The confident pseudo-labels have a smaller KL divergence on average compared to the labeled data (1.69 vs. 2.70), yet its distribution shift is still significant. Interestingly, the average KL divergence of all the pseudo-labels is much smaller than that of the other two groups (0.45 vs. 1.69/2.70). In other words, selecting confident pseudo-labels reduces the errors from noisy pseudo-labels, but also introduces data bias due to the biased selection. Additionally, it is not easy to decide a proper threshold for sample selection, especially for the fine-grained localization task that is more sensitive to noisy labels. Moreover, coordinate-regression based methods usually output predictions without confidence scores, which makes classic self-training inapplicable. Thus, we pose the natural question: \textit{Is pseudo-label selection a necessary component of self-training?}  

\begin{figure}[t]
\centering
  \includegraphics[width=1\linewidth]{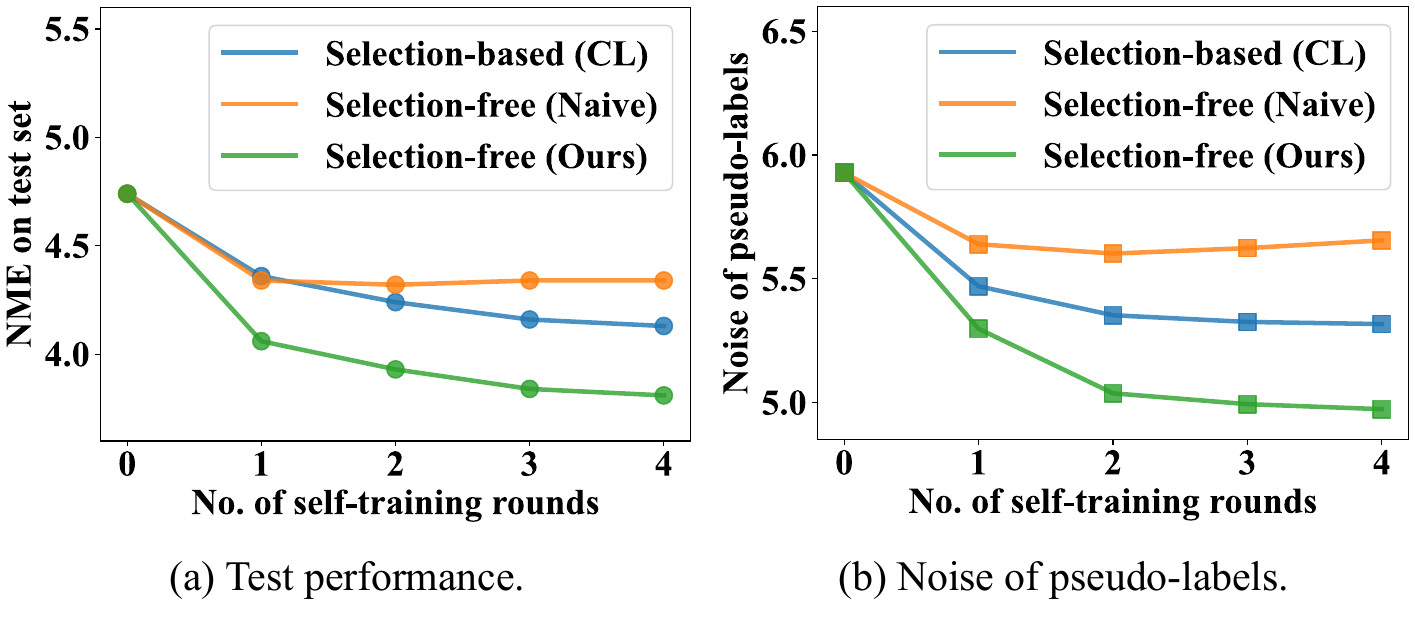}
\caption{Comparison of selection-based and selection-free methods over the rounds, trained on 300W~\cite{STZ13} with 5\% labeled. (a) Compare on test performance. (b) Compare on the noise of the estimated pseudo-labels.} 
\label{fig:intro_compare}      
\end{figure}

\subsection{Selection-free Self-training}
\label{sec3.3}

To answer the question above, we first reformulate self-training as a selection-free method: 
\begin{equation}
\mathcal{R}^{\text{SFST}}(f_{\bm{\theta}}) = \mathcal{R}^{\text{SL}}(f_{\bm{\theta}}) + \lambda \cdot \sum_{\mathclap{ (\bm{x}_i,\bm{y}_i) \in (\mathcal{X}_u, \widehat{\mathcal{Y}}_u) }} \mathcal{G}(f_{\bm{\theta}}(\bm{x}_i), {\bm{y}_i}),
\label{eqn4}
\end{equation}
where $\mathcal{G}$ is a regularization term that can be any tasks or techniques. Compared to Eqn.~\ref{eqn3}, the differences are mainly two-fold: 1) The selection-free method uses all the pseudo-labels in each round; 2) the leveraging of pseudo-labels is not limited to the standard landmark detection task. Accordingly, the core of selection-free self-training is to design appropriate instantiations of $\mathcal{G}$ that can successfully leverage the pseudo-labels while taking care of noise.

Before exploring more sophisticated designs, a naive strategy can be easily obtained by setting $\mathcal{G}$ to standard landmark detection (equivalent to selection-based self-training with $\tau=0$), which serves as a baseline. To understand how the naive method performs, we compare it to the recently proposed selection-based method CL~\cite{BTQ21}, in terms of test performance and pseudo-label noise. Fig.~\ref{fig:intro_compare}a and Fig.~\ref{fig:intro_compare}b show the normalized mean error (NME) on the 300W test set and the noise of the predicted pseudo-labels (i.e., the difference between the pseudo-labels and GTs), respectively. We can see that the naive selection-free method is inferior to CL~\cite{BTQ21} in terms of test NME. Specifically, the naive method quickly converges to a poor performance and its pseudo-label noise even increases after two rounds, which we believe is due to severe confirmation bias. In contrast, the test NME as well as the pseudo noise of CL gradually decreases over the rounds. We attribute the success of CL to its data-level curriculum, where confident pseudo-labels are utilized first, to ease the confirmation bias. Therefore, we believe the key to self-training is to utilize reliable information first, while \textit{obtaining such information is not necessarily restricted to the selection of confident pseudo-labels}. For example, it is also possible to extract reliable information by conducting a simpler but confident task on pseudo-labels.

\section{Self-training for Landmark Detection}
\label{sec4}

Based on the above observation, we identify that the key to successful selection-free self-training is to construct a task curriculum, where confident tasks are conducted first on pseudo-labels to provide reliable information. To this end, we propose \textit{pseudo pretraining} and \textit{shrink regression} as two essential components of the curriculum, which will be introduced in Sec.~\ref{sec4.1} and~\ref{sec4.2}, respectively. The overall training pipeline is given in Sec.~\ref{sec4.3}.

\begin{figure}[t]
\centering
  \includegraphics[width=0.85\linewidth]{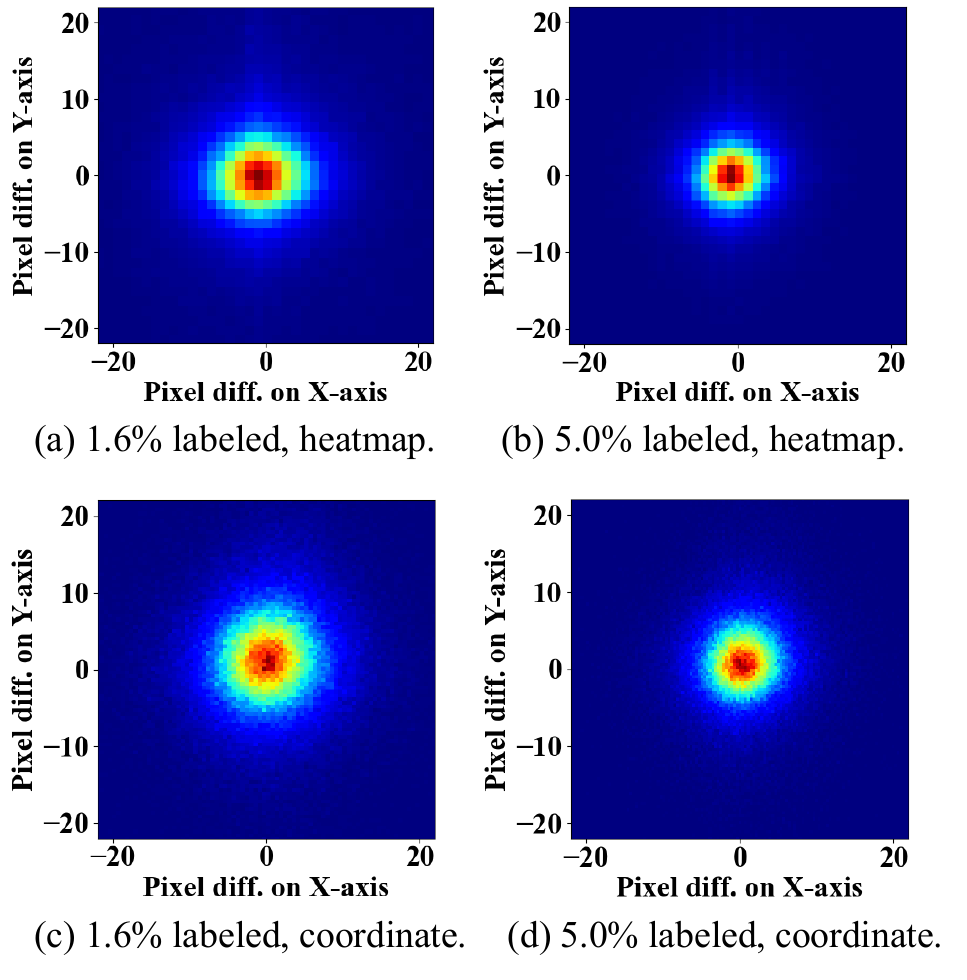}
\caption{2D histograms of the offsets of pseudo-labels relative to GTs, trained on 300W with different labeled ratios. We analyzed both heatmap ((a)-(b)) and coordinate ((c)-(d)) models.} 
\label{fig:noise_dist}      
\end{figure}

\subsection{Pseudo Pretraining for Better Initialization}
\label{sec4.1}

During the pseudo pretraining stage, i.e., the simplest task of the curriculum, a model is not asked to fit the pseudo-labels directly, but rather obtaining a better initialization with them. To this end, the model initialized from ImageNet~\cite{DDS09} pretrained weights is trained on \textit{all} the pseudo-labeled data with the following formula: 
\begin{equation}
\bm{\theta}_{\text{pre}} = \arg \min_{\bm{\theta}} \sum_{\mathclap{ (\bm{x}_i,\bm{y}_i) \in (\mathcal{X}_u, \widehat{\mathcal{Y}}_u) }} \mathcal{L} (f_{\bm{\theta}}(\bm{x}_i), {\bm{y}_i}), 
\label{eqn5}
\end{equation}
where $\mathcal{L}$ is the same loss as in supervised landmark detection. After obtaining a better initialization with pseudo pretraining, the model continues with the second training stage, which will be described in Sec.~\ref{sec4.3}. Despite its simplicity, the design of pseudo pretraining is able to leverage all the pseudo-labels while handling the noise in an adaptive way. An analysis has been performed to verify this (see Sec.~\ref{sec5.4}), which shows that the noisier pseudo-labels learned during the pseudo pretraining stage are more likely to be forgotten after the second training stage. Intuitively speaking, when the model learns incorrect information from the pseudo-labels at the pretraining stage, it will gradually forget it at the second training stage because the incorrect information would be inconsistent with the `correct' labels.

\begin{figure}[t]
\centering
  \includegraphics[width=0.7\linewidth]{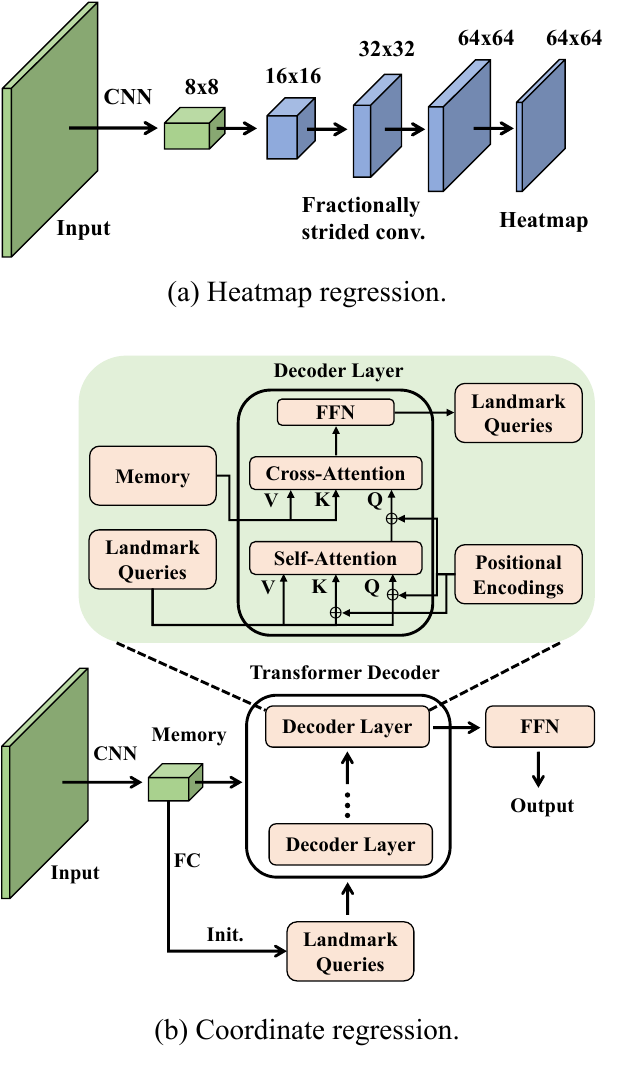}
\caption{Architecture of the two base models for landmark detection. (a) Heatmap regression. (b) Coordinate regression with transformer decoder as the head.} 
\label{fig:framework_hm_tf}      
\end{figure}

\subsection{Shrink Regression for Noise Resistance}
\label{sec4.2}

Although pseudo pretraining handles noisy samples well, it has not fully leveraged the pseudo-labels as they are only used for model initialization. To achieve better performance, it is necessary to utilize pseudo-labels directly for model training while taking care of the noise. To this end, we first inspect the characteristics of noisy labels by visualizing the offsets of pseudo-labels relative to the GTs in 2D histograms, as shown in Fig.~\ref{fig:noise_dist}. The figure indicates that pseudo-labels are distributed around GTs for both heatmap (Figs.~\ref{fig:noise_dist}a and~\ref{fig:noise_dist}b) and coordinate regression (Figs.~\ref{fig:noise_dist}c and~\ref{fig:noise_dist}d) across different labeled ratios. For those pseudo-labels that are relatively far from the GTs, if the model directly fits to them with the standard granularity\footnote{The granularity here refers to the precision of regression task. A fine-grained regression task (i.e., high-level granularity) penalizes more on small prediction errors while a coarse-grained regression (i.e., low-level granularity) is more tolerant to small errors.}, it would learn the noisy information introduced by these pseudo-labels due to their deviation from GTs. On the other hand, if a coarse granularity is used for regression, the model fits to a broader region centered on the pseudo-labels, which is more likely to cover the GTs. Accordingly, we argue that \textit{the standard regression task on pseudo-labels can be made more confident by reducing the granularity of regression targets}. In other words, it sacrifies the precision of regression for the reliability of learning targets. Following this observation, we propose shrink regression, which starts with coarse regression, then progressively increases the granularity until the same level as standard regression. When it is applied, the $\mathcal{G}$ in Eqn.~\ref{eqn4} can be implemented as the loss function $\mathcal{L}^{\text{SR}}(f_{\bm{\theta}}(\bm{x}), \bm{y}, t)$, where $t$ represents the $t$-th round and $t=2,...,T$. Here $t$ starts from 2 because shrink regression is applied from the second round. Different instantiations of $\mathcal{L}^{\text{SR}}$ for heatmap and coordinate regression are detailed in Sec.~\ref{sec4.2.1} and Sec.~\ref{sec4.2.2}, respectively. A supervised landmark detection model can be seen as a special case of shrink regression, where its granularity is irrelevant to $t$ and fixed to the standard value. 

\begin{figure}[t]
\centering
  \includegraphics[width=0.71\linewidth]{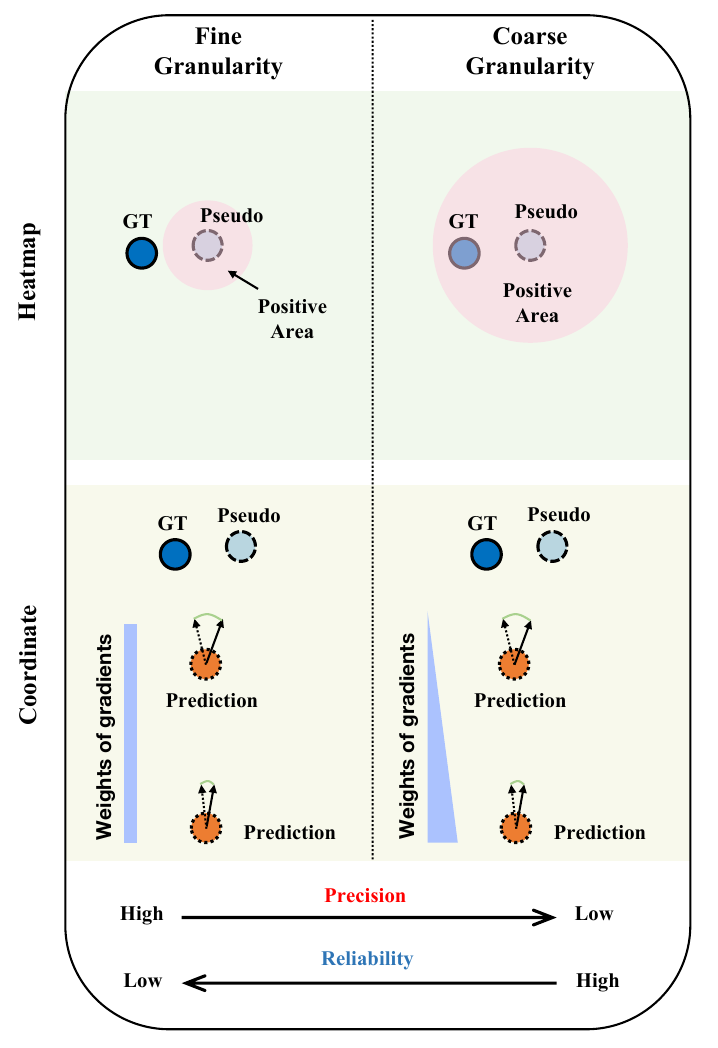}
\caption{A schematic diagram that shows the precision-reliability trade-off of heatmap (upper) and coordinate regression (lower).} 
\label{fig:shrink_diagram}     
\end{figure}

\subsubsection{Heatmap Regression}
\label{sec4.2.1}

Fig.~\ref{fig:framework_hm_tf}a gives the architecture of the base heatmap regression model (denoted as HM), where it predicts landmarks through heatmaps. The input size is $256\times256$. After obtaining a feature map with a convolutional neural network (CNN), it further goes through three fractionally strided convolutional layers to increase the map resolution, and finally outputs a heatmap with a convolutional layer. During inference, the coordinates of landmarks can be obtained with a post-processing step, which is detailed in~\cite{XWW18}.

When shrink regression is applied, we adjust the granularity of heatmap regression by adjusting its training targets. Specifically, heatmap regression uses a proxy heatmap $\bm{m} \in \mathbb{R}^{N \times H \times W}$ for training and maps $\bm{y}$ to 2D Gaussians with centers $\bm{\mu}$ and standard deviation $\sigma$ through a mapping $\phi$. By increasing $\sigma$, the positive areas in $\bm{m}$ are enlarged, thus becoming coarser and more tolerant to small prediction errors. Therefore, the shrink regression for heatmap regression can be formulated as 
\begin{equation}
\begin{split}
\mathcal{L}^{\text{SR}}_{\text{map}}(f_{\bm{\theta}}(\bm{x}), \bm{y}, t) & = \mathcal{L} (f_{\bm{\theta}}(\bm{x}), \phi(\bm{y}, t)) \\
& = \mathcal{L} (f_{\bm{\theta}}(\bm{x}), \bm{\mu}, \sigma_t), 
\end{split}
\label{eqn6}
\end{equation}
where $\mathcal{L}$ is a standard loss function for heatmap regression (e.g., $\mathcal{L}_2$ loss), and $\sigma_t$ is the standard deviation at the $t$-th round. To construct a task curriculum that transitions from coarse to fine, $\sigma_{t+1}$ will be smaller than $\sigma_t$, and $\sigma_T=\sigma_{\text{std}}$, where $\sigma_{\text{std}}$ is the standard value for supervised learning and $T$ represents the last self-training round. 

We give a schematic diagram to show why reducing regression granularity can increase the reliability of learning targets, which is shown in Fig.~\ref{fig:shrink_diagram}. We can see from the figure that when a heatmap regression model is trained with pseudo-labels in fine granularity (upper left of the figure), the training target may not reflect the real location (i.e., GT) due to the noise from pseudo-labels. On the other hand, when trained with coarse granularity (upper right of the figure), the enlarged positive area is more likely to contain GT, which provides more reliable learning targets.  

\subsubsection{Coordinate Regression}
\label{sec4.2.2}

We adopt the transformer-based model from~\cite{JLL21} as our coordinate regression model, denoted as TF. Fig.~\ref{fig:framework_hm_tf}b gives the architecture of the TF model. TF has the same CNN backbone as that of HM, but it uses a transformer decoder as the detection head. The head takes as inputs $N$ landmark queries and a memory extracted with CNN, where $N$ is the number of landmarks. The landmark queries are initialized with the predictions directly from the memory via a fully connected layer, and then goes through multiple decoder layers for query refinement. Lastly, the landmark queries will be converted to coordinates (normalized to [0, 1]) via a feed forward network (FFN). 

\begin{figure}[t]
\centering
  \includegraphics[width=1\linewidth]{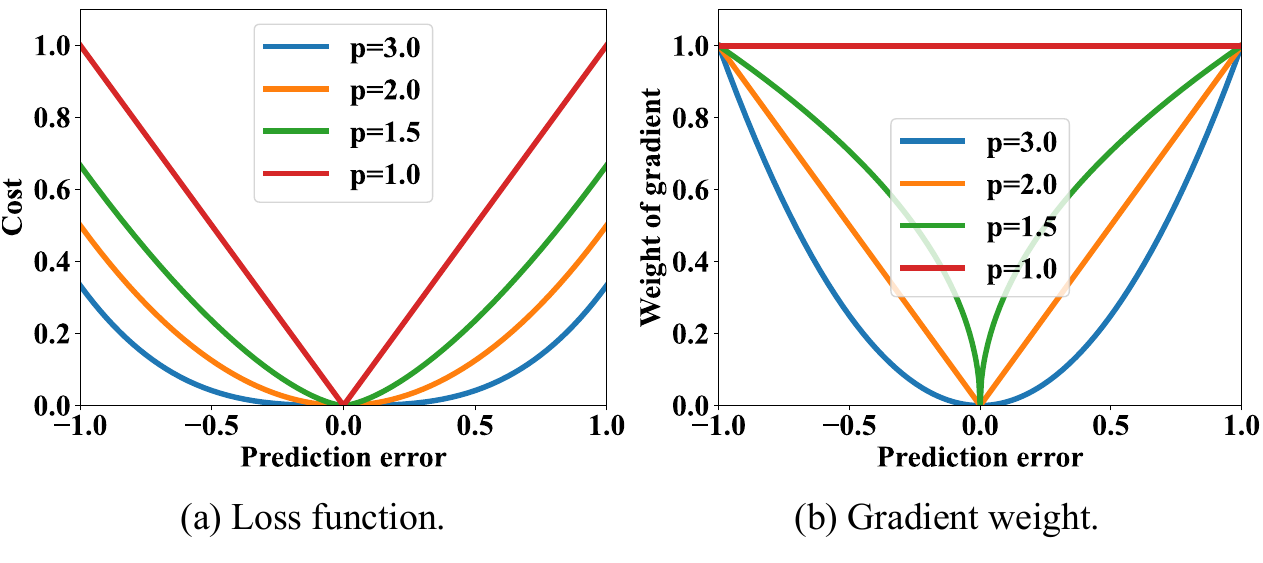}
\caption{Visualization of shrink loss and its gradient weight with different power $p$. (a) The loss function. (b) The gradient weight.} 
\label{fig:shrink_loss}     
\end{figure}

Since coordinate regression is end-to-end optimized w.r.t. the coordinates, we adjust its granularity via the loss function. In supervised learning, $\mathcal{L}_1$ loss is widely used for coordinate-based models since it has been shown to be superior to $\mathcal{L}_2$ loss~\cite{FKA18}. Formally, $\mathcal{L}_1$ and $\mathcal{L}_2$ are defined as
\begin{equation}
\mathcal{L}_1(\hat{\bm{y}}, \bm{y}) = |\hat{\bm{y}} - \bm{y}| \quad \text{and} \quad \mathcal{L}_2(\hat{\bm{y}}, \bm{y}) = \frac{1}{2}(|\hat{\bm{y}} - \bm{y}|)^2,
\label{eqn7}
\vspace{-4pt}
\end{equation}
where $\hat{\bm{y}}$ and $\bm{y}$ are the prediction and GT, respectively. We further write the gradients of the two losses w.r.t. $\bm{\theta}$ as
\begin{equation}
\frac{\partial \mathcal{L}_1(\hat{\bm{y}}, \bm{y})}{\partial \bm{\theta}} = \pm 1 \cdot \nabla_{\bm{\theta}} \hat{\bm{y}} \quad \text{and}
\label{eqn8}
\end{equation}
\begin{equation}
\frac{\partial \mathcal{L}_2(\hat{\bm{y}}, \bm{y})}{\partial \bm{\theta}} = \pm |\hat{\bm{y}} - \bm{y}| \cdot \nabla_{\bm{\theta}} \hat{\bm{y}}.
\label{eqn9}
\end{equation}
From Eqn.~\ref{eqn8} and~\ref{eqn9}, we can see that the weight of the gradient becomes small when the prediction error $|\hat{\bm{y}} - \bm{y}|$ becomes small in the $\mathcal{L}_2$ loss, while the weight in the $\mathcal{L}_1$ loss always equals 1. This may explain why $\mathcal{L}_1$ is preferred to $\mathcal{L}_2$ for supervised learning, as the former provides more precise results by focusing more on small errors. However, we argue, based on the observation in Sec.~\ref{sec4.2}, that $\mathcal{L}_2$ is more robust to noisy labels (within [-1, 1]) because it downweights the gradients of small errors. That is to say, $\mathcal{L}_2$ loss has a coarser granularity than $\mathcal{L}_1$. To make the loss adjustable in terms of granularity, we first generalize it to $\mathcal{L}_p$ loss, which is formulated as 
\begin{equation}
\mathcal{L}_{p}(\hat{\bm{y}}, \bm{y}) = \frac{1}{p}(|\hat{\bm{y}} - \bm{y}|)^{p}, \quad p \geq 1,
\label{eqn10}
\end{equation}
where $p$ can be adjusted to change the granularity. Fig.~\ref{fig:shrink_loss} visualizes the $\mathcal{L}_p$ loss and its gradient weights with varying $p$. As can be seen from Fig.~\ref{fig:shrink_loss}b, a larger $p$ has smaller gradient weights on small prediction errors; thus it is coarser and more tolerant to small errors. With the $\mathcal{L}_p$ loss, we can obtain shrink loss by shrinking $\mathcal{L}_p$ over $t$:
\begin{equation}
\mathcal{L}^{\text{SR}}_{\text{coord}}(f_{\bm{\theta}}(\bm{x}), \bm{y}, t) = \mathcal{L}_{p_t} (f_{\bm{\theta}}(\bm{x}), \bm{y}), 
\label{eqn11}
\end{equation}  
where $p_t$ is the norm of the loss at the $t$-th round. Similar to heatmap rgression, $p_{t+1}$ should be smaller than $p_t$, and $p_T=1$ (i.e., using $\mathcal{L}_1$ loss in the last round). The formulation of the shrink loss is also supported by our analysis of gradient correlations between the pseudo-labels and GTs (see Sec.~\ref{sec5.4}), showing that the gradients of small errors from pseudo-labels are indeed noisier.

We use the schematic diagram in Fig.~\ref{fig:shrink_diagram} to demonstrate why the weights of gradients are related to the reliability of learning. Coordinate regression with $\mathcal{L}_1$ loss assigns the same weight for all the gradients (lower left of the figure), regardless of whether the prediction error is large or small. However, as shown in the figure, when the prediction is closer to the pseudo-label, it is more likely to yield a wrong gradient due to the deviation between the GT and pseudo-label. Therefore, the gradient with larger prediction errors are more reliable, which motivates us to downweight the gradients of small prediction errors for improving the learning reliability (lower right of the figure).    

\begin{algorithm}[t]
\caption{Self-training for Landmark Detection}
\begin{algorithmic}[1]
\Require Labeled data ($\mathcal{X}_l$, $\mathcal{Y}_l$) and unlabeled data $\mathcal{X}_u$.
\Ensure Model parameters $\bm{\theta}$.
\State Train a model on ($\mathcal{X}_l$, $\mathcal{Y}_l$) w/ supervised learning.
\State Estimate $\widehat{\mathcal{Y}}_u$ for $\mathcal{X}_u$ w/ the trained model.
\For{$t \gets$ $1$ to \textit{T}} 
\State Train a new model on ($\mathcal{X}_u$, $\widehat{\mathcal{Y}}_u$) w/ pseudo pretraining.
\If {$t = 1$}
\State Continue training on ($\mathcal{X}_l$, $\mathcal{Y}_l$) w/ standard regression.
\Else
\State Continue training on ($\mathcal{X}_l$, $\mathcal{Y}_l$) and ($\mathcal{X}_u$, $\widehat{\mathcal{Y}}_u$) w/ standard and shrink regression, respectively.
\EndIf
\State Estimate and update $\widehat{\mathcal{Y}}_u$ w/ the trained model.
\EndFor
\end{algorithmic}
\label{alg1}
\end{algorithm}

\subsection{Training Pipeline}
\label{sec4.3}

Similar to selection-based self-training, STLD alternately trains with labeled set and estimates pseudo-labels for unlabeled images, but differs in how to utilize pseudo-labels. Instead of adding confident pseudo-labels to the labeled set, STLD utilizes all of them via a task curriculum. To be specific, each round of STLD consists of two stages, where the first stage conducts pseudo pretraining on all the pseudo-labeled data and the second stage directly leverages the labeled set via source-aware mixed training. The mixed training in the first round only conducts standard regression on GTs. From the second round, shrink regression is further added to the mixed training for leveraging the pseudo-labels, and the objective function at this stage is defined as 
\begin{equation*}
\begin{split}
\mathcal{R}^{\text{STLD}}(f_{\bm{\theta}}) = \mathcal{R}^{\text{SL}}(f_{\bm{\theta}_{\text{pre}}}) &+ \lambda \cdot 
\sum_{\mathclap{ (\bm{x}_i,\bm{y}_i) \in \mathcal{B}}} \mathcal{L}^{\text{SR}} (f_{\bm{\theta}_{\text{pre}}}(\bm{x}_i), {\bm{y}_i}, t)
, \\
& \mathcal{B}= \left\{
\begin{array}{ll}
      \varnothing & t = 1 \\
      (\mathcal{X}_u, \widehat{\mathcal{Y}}_u) & t \geq 2 \\
\end{array} 
\right.,
\end{split}
\label{eqn12}
\end{equation*}
where $\theta_{\text{pre}}$ indicates model initialization from the first stage and $\lambda$ is a balancing coefficient. We set $\lambda=0.1$ when $t<T$, and $\lambda=1$ when $t=T$ because shrink regression becomes standard regression in the last round. The training pipeline of STLD is given in Alg.~\ref{alg1}. The analysis in Fig.~\ref{fig:intro_compare} shows that STLD improves over the naive method significantly and outperforms CL by a large margin as it is able to fully leverage the pseudo-labels while handling their noise appropriately. 

\noindent \textbf{Speed up training.}
It can be time consuming to conduct pseudo pretraining from scratch in each round. Empirically, we found the training can be accelerated by initializing with the pseudo pretrained model from the last round and reducing the number of training epochs to 1/5 of the original. In this way, only the first round will conduct a complete pseudo pretraining, which largely saves the training time. 

\begin{table*}[t]
\centering
\small
\newcolumntype{C}{>{\centering\arraybackslash}X}%
\caption{Comparison with existing semi-supervised landmark detection methods on 300W, WFLW, and AFLW. Results are in NME (\%). $*$ indicates usage of external unlabeled data. $\dagger$ indicates our implementation. The best results are in \textbf{bold} and the second best are \underline{underlined}.}
\begin{tabularx}{\linewidth}{lCCCCCCCCCCCC}
\toprule	 
Dataset & \multicolumn{4}{c}{300W} & \multicolumn{4}{c}{WFLW} & \multicolumn{4}{c}{AFLW}\\ \cmidrule(r){2-5} \cmidrule(r){6-9} \cmidrule(r){10-13}
Labeled ratio  & 1.6\% & 5\% & 10\% & 20\% & 0.7\% & 5\% & 10\% & 20\% & 1.0\% & 5\% & 10\% & 20\%\\
\midrule	
RCN+~\cite{HMT18} & - & 5.11 & 4.47 & 4.14 & - & - & - & - & 2.88 & 2.17 & - & -\\
TS$^3$~\cite{DoY19}  & - & - & 5.64 & 5.03 & - & - & - & - & - & 2.19 & 2.14 & 1.99\\
3FabRec$^*$~\cite{BjC20} & 5.10 & 4.75 & 4.47 & 4.31 & 8.39 & 7.68 & 6.73 & 6.51 & 2.38 & 2.13 & 2.03 & 1.96\\
STC~\cite{JLS21} & 4.72 & 4.04 & 3.76 & 3.58 & 7.99 & 5.94 & 5.40 & 5.09 & \underline{2.05} & 1.79 & 1.74 & 1.68\\
PL$^{\dagger}$~\cite{Lee13} & 5.45 & 4.33 & 3.80 & 3.60 & 9.04 & 6.00 & 5.47 & 5.14 & 2.20 & 1.85 & 1.78 & 1.70\\
Data Dist.$^{\dagger}$~\cite{RDG18} & 5.38 & 4.35 & 3.82 & 3.59 & 8.99 & 6.06 & 5.50 & 5.18 & 2.18 & 1.86 & 1.80 & 1.72\\
CL$^{\dagger}$~\cite{BTQ21} & 5.61 & 4.16 & 3.76 & 3.58 & 8.87 & 5.98 & 5.45 & 5.14 & 2.13 & 1.83 & 1.75 & 1.70\\
Easy-hard Aug.$^{\dagger}$~\cite{XWZ21} & 5.00 & 4.18 & 3.78 & 3.60 & 8.63 & 5.90 & 5.41 & 5.14 & 2.10 & 1.80 & 1.75 & 1.69\\
FixMatch$^{\dagger}$~\cite{SBL20} & 4.94 & 4.15 & 3.76 & 3.59 & 8.57 & 5.86 & 5.39 & 5.12 & 2.09 & 1.79 & 1.74 & 1.69\\
DST$^{\dagger}$~\cite{CJW22} & 4.66 & 4.02 & 3.75 & 3.57 & 7.95 & 5.81 & 5.36 & 5.10 & 2.06 & 1.78 & 1.73 & 1.67 \\
HybridMatch~\cite{KLK23} & \underline{3.76} & \underline{3.50} & 3.44 & 3.40 & \textbf{5.47} & \underline{5.04} & 4.85 & 4.73 & \textbf{1.77} & \underline{1.68} & 1.66 & 1.61 \\
\midrule	
STLD-HM-R18 (Ours) & 4.47 & 3.81 & 3.64 & 3.52 & 7.91 & 5.67 & 5.30 & 5.06 & 2.03 & 1.75 & 1.69 & 1.63\\
STLD-TF-R18 (Ours) & 4.97 & 4.00 & 3.69 & 3.49 & 8.07 & 5.68 & 5.17 & 4.82 & 2.10 & 1.76 & 1.67 & 1.60\\
STLD-HM-HRNet (Ours) & \textbf{3.73} & \textbf{3.40} & \textbf{3.32} & \underline{3.27} & \underline{6.09} & \textbf{5.02} & \underline{4.79} & \underline{4.58} & \underline{1.87} & \textbf{1.65} & \underline{1.61} & \underline{1.55} \\ 
STLD-TF-HRNet (Ours) & 4.35 & 3.69 & \underline{3.42} & \textbf{3.25} & 6.65 & 5.06 & \textbf{4.77} & \textbf{4.46} & 1.96 & \textbf{1.65} & \textbf{1.60} & \textbf{1.53} \\ 
\bottomrule
\end{tabularx}
\label{tab:results_semi}
\end{table*}

\section{Experiments}
\label{sec5}

We introduce datasets and settings in Sec.~\ref{sec5.1} and~\ref{sec5.2}, respectively, then compare STLD with SOTAs in Sec.~\ref{sec5.3}. Lastly, we perform model analysis in Sec.~\ref{sec5.4}. 

\subsection{Datasets}
\label{sec5.1}

\subsubsection{Facial landmark} 
\textbf{300W}~\cite{STZ13} is a popular benchmark for facial landmark detection. There are 3,148 training images and 689 test images in total, where each image contains 68 landmarks. In this work, we report results on the full test set. \textbf{WFLW}~\cite{WQY18} is another widely used benchmark, which contains 7,500 images for training and 2,500 for testing. The images were collected from WIDER Face~\cite{YLL16}, and each of them was annotated with 98 landmarks. \textbf{AFLW}~\cite{KWR11} is a large-scale benchmark containing both profile and frontal faces. The training set includes 20,000 images and the test set has 4,386. We use the 19 annotated landmarks of AFLW, following~\cite{ZLL16a}. For the preprocessing of the three datasets, we crop faces with the provided bounding boxes, then resize to $256 \times 256$. Additionally, we use the 200K images from CelebA~\cite{LLW15} as the extra unlabeled data in the omni setting \textit{only}.

\subsubsection{Medical landmark}
The \textbf{Hand} is a public dataset\footnote{https://ipilab.usc.edu/research/baaweb} of hand X-ray images. There are 609 and 300 images for training and testing, respectively. We resize each image to $512 \times 512$ for preprocessing. Payer et al.~\cite{PSB19} labeled 37 landmarks for each image and assume 50mm wrist width as physical distance, which is adopted in this paper. Additionally, we use 12K images from RSNA~\cite{HPC19} training set as the extra unlabeled data for omni-supervised learning.

\begin{figure}[t]
\centering
  \includegraphics[width=1\linewidth]{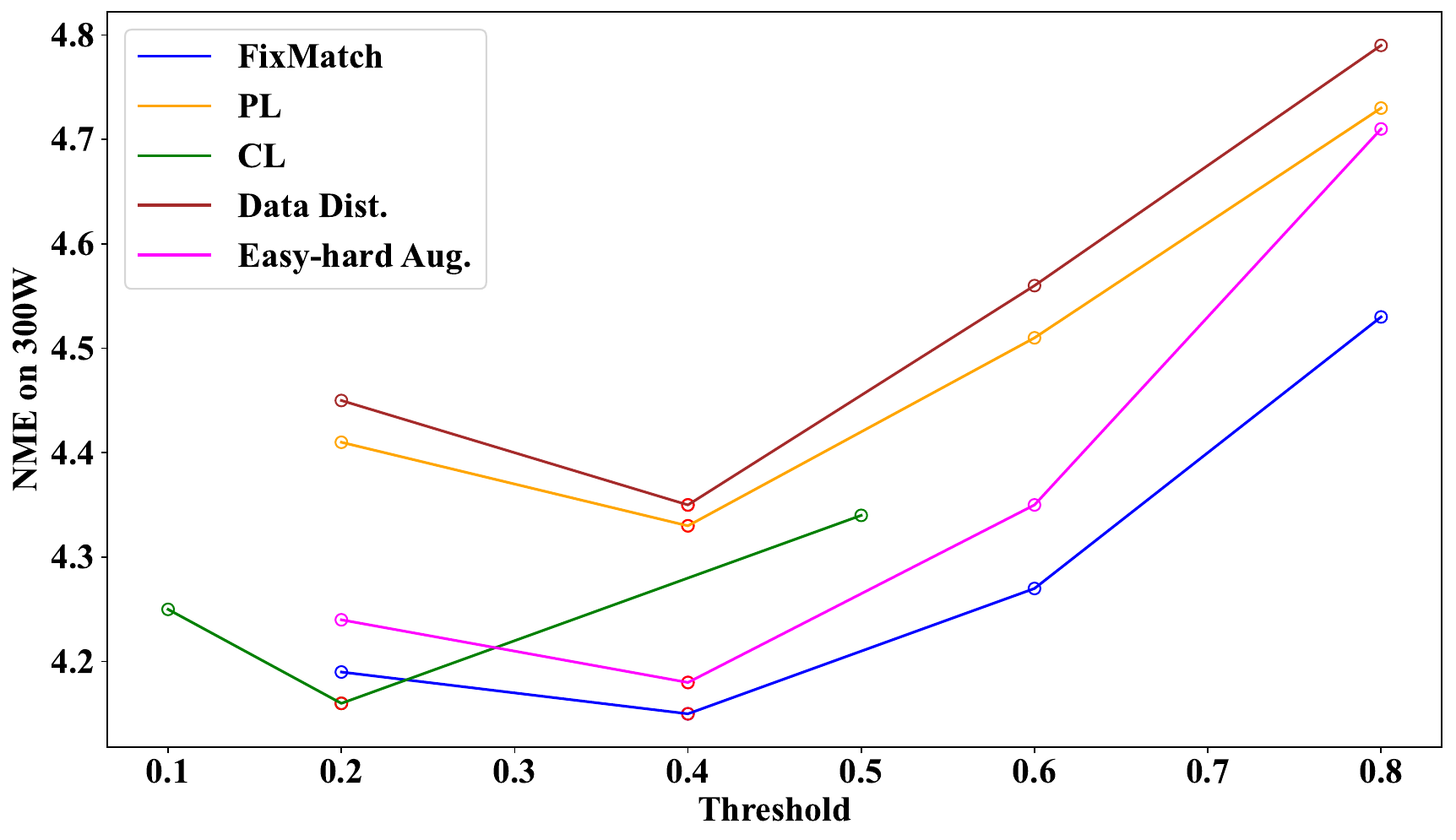}
\caption{Performance of selection-based methods with different thresholds. The setting is under 300W with 5\% labeled data. The threshold for CL refers to the curriculum step at each round. \textcolor{red}{A red dot} indicates the optimal threshold for each method, which was used in the experiments.} 
\label{fig:selection_thresh}      
\end{figure}

\subsection{Experimental Settings}
\label{sec5.2}

\subsubsection{Implementation details}
To demonstrate the universality of our method, we use two base models, one is heatmap-based and the other is coordinate-based. The heatmap model (HM) is adopted from~\cite{XWW18} and the coordinate model uses transformer decoder as the head (TF), adopted from~\cite{JLL21}. ResNet-18~\cite{HZR16} is used as the default backbone and HRNet~\cite{WSC19} is also used to explore better performance. Both networks are initialized from ImageNet~\cite{DDS09} pretrained weights. The initial learning rate is 1e-4, and Adam~\cite{KiB15} is used as the optimizer. The batch size is set to 16. At each training stage, the total training epoch is 60 for HM, learning rate decayed by 10 at 40th and 50th epoch; the training epoch is 360 for TF, decayed by 10 at 240th and 320th epoch. For standard regression, HM uses $\sigma=1.5$ and TF uses $p=1$ (i.e., $\mathcal{L}_1$ loss), following common practice~\cite{WSC19,JLL21}. The code was implemented with PyTorch 1.13. 

\subsubsection{Semi-supervised setting}
For the labeled ratios of each dataset, we follow the previous works~\cite{HMT18,DoY19,BjC20} by adopting the most common ratios and we randomly split the original training set into labeled and unlabeled sets. The number of self-training rounds is set to 4. For the granularity curriculum of shrink regression, HM uses [2.2, 1.8, 1.5], and TF uses [2.4, 1.6, 1.0]. In Sec.~\ref{sec5.4}, we show that shrink regression is relatively robust to different curriculum settings. 

\subsubsection{Omni-supervised setting}
The omni setting was proposed for a more practical scenario~\cite{RDG18}. As a special regime of SSL, it uses all available labeled data together with large-scale unlabeled data for model training. We do experiments in the omni setting on 300W~\cite{STZ13} and Hand, where CelebA~\cite{LLW15} and RSNA~\cite{HPC19} are used as the extra unlabeled images for facial and medical data, respectively.

\subsubsection{Evaluation metric}
Following~\cite{WSC19}, we use normalized mean error (NME), area-under-the-curve (AUC), and failure rate (FR) at 10\% as the metrics for facial landmark data. For NME, 300W and WFLW use inter-ocular distance for normalization and AFLW uses image size. For the medical data, we use mean radial error (MRE), following~\cite{ZYX21}. 

\begin{table}[t]
\centering
\small
\newcolumntype{C}{>{\centering\arraybackslash}X}%
\caption{AUC and FR$_{10\%}$ on 300W test set, with 10\% and 20\% labeled data. $\dagger$ indicates our implementation. The best results are in \textbf{bold}.}
\begin{tabularx}{\linewidth}{lCCCC}
\toprule	 
\multirow{2}{*}{Method} & \multicolumn{2}{c}{10\%} & \multicolumn{2}{c}{20\%} \\ \cmidrule(r){2-3} \cmidrule(r){4-5}
 & AUC$\uparrow$ & FR$\downarrow$ & AUC$\uparrow$ & FR$\downarrow$ \\
\midrule
CL$^{\dagger}$~\cite{BTQ21} & 62.26 & 0.73 & 64.46 & 0.44 \\
FixMatch$^{\dagger}$~\cite{SBL20} & 62.34 & 0.87 & 64.06 & 0.87 \\
DST$^{\dagger}$~\cite{CJW22} & 62.58 & 0.44 & 64.13 & 0.29 \\
HybridMatch~\cite{KLK23} & - & - & 60.56 & 0.17 \\
\midrule
STLD-HM-R18 (Ours) & 63.30 & 0.44 & 65.26 & 0.29 \\
STLD-TF-R18 (Ours) & 62.79 & 0.58 & 65.26 & 0.29 \\
STLD-HM-HRNet (Ours) & \textbf{66.63} & \textbf{0.29} & \textbf{67.74} & \textbf{0.15} \\
STLD-TF-HRNet (Ours) & 65.64 & \textbf{0.29} & 67.44 & \textbf{0.15} \\
\bottomrule
\end{tabularx}
\label{tab:results_auc_fr}
\end{table} 

\begin{table}[t]
\centering
\small
\newcolumntype{C}{>{\centering\arraybackslash}X}%
\caption{Comparison with existing SSL methods on 300W in the omni setting. $\dagger$ indicates our implementation. The best results are in \textbf{bold} and the second best are \underline{underlined}.}
\begin{tabularx}{\linewidth}{llCC}
\toprule	 
Method & Backbone & Unlabeled & NME (\%) \\ 
\midrule	
TS$^3$~\cite{DoY19} & HG+CPM & 20K & 3.49\\
LaplaceKL~\cite{RLZ19} & - & 70K & 3.91\\
DeCaFA~\cite{DBC19} & U-net & 200K & 3.39\\
STC~\cite{JLS21} & ResNet-18 & 200K & 3.23\\
Data Dist.$^{\dagger}$~\cite{RDG18} & ResNet-18 & 200K & 3.31\\
CL$^{\dagger}$~\cite{BTQ21} & ResNet-18 & 200K & 3.26\\
Easy-hard Aug.$^{\dagger}$~\cite{XWZ21} & ResNst-18 & 200K & 3.27\\
\midrule	
STLD-HM (Ours) & ResNet-18 & 20K & 3.29 \\
& & 70K & 3.27 \\
& & 200K & 3.23 \\
STLD-TF (Ours) & ResNet-18 & 20K & 3.20 \\
& & 70K & \underline{3.18} \\
& & 200K & \textbf{3.14} \\
\bottomrule
\end{tabularx}
\label{tab:results_omni_300w}
\end{table}

\subsection{Results}
\label{sec5.3}

\subsubsection{Semi-supervised Setting}
\label{sec5.3.1}

In addition to the existing semi-supervised landmark detection models~\cite{HMT18,DoY19,BjC20,JLS21,KLK23}, we also compare with semi-supervised pose estimation methods~\cite{RDG18,XWZ21} and generic self-training methods~\cite{Lee13,BTQ21,SBL20,CJW22}. For selection-based methods, we select their optimal thresholds based on the performance on 300W with 5\% labeled data, which can be found in Fig.~\ref{fig:selection_thresh}. Following~\cite{KLK23}, we also equip our model with HRNet~\cite{WSC19} for better performance. Tab.~\ref{tab:results_semi} shows the results on 300W, WFLW, and AFLW with different labeled ratios. We can see that STLD obtains the best results on most of the entries. Specifically, STLD-HM-R18 and STLD-TF-R18 both outperform their ResNet-18 counterparts on almost all the entries, demonstrating the effectiveness of the proposed method. When a heavier backbone such as HRNet is equipped, the performance of STLD is further boosted. For example, STLD-HM-HRNet achieves 3.73, 3.40, and 3.32 of NME (\%) on 300W for 1.6\%, 5\%, and 10\% labeled data, respectively, outperforming all the previous methods. On WFLW and AFLW, STLD-HM-HRNet achieves the best results with 5\% labeled data. In contrast, STLD-TF-HRNet performs better with higher labeled ratios, thanks to the representation capability of transformers. For example, it obtains 3.25 of NME (\%) on 300W and 1.53 on AFLW with 20\% labeled, both of which are slightly better than STLD-HM-HRNet. The advantage of STLD-TF-HRNet is even larger on WFLW, achieving 4.46 with 20\% labeled, which is about 5.7\% improvement over the best existing method HybridMatch (4.73). To make the evaluation comprehensive, we also include the results under AUC and FR$_{10\%}$, which are shown in Tab.~\ref{tab:results_auc_fr}. Again, our STLD with HRNet obtains the SOTA performance on both metrics, validating the superiority of the proposed method.

\begin{table}[t]
\centering
\small
\newcolumntype{C}{>{\centering\arraybackslash}X}%
\caption{Comparison with existing supervised learning methods on Hand in the omni setting. $\dagger$ indicates our implementation. The best results are in \textbf{bold} and the second best are \underline{underlined}.}
\begin{tabularx}{\linewidth}{llCC}
\toprule	 
Method & Backbone & Unlabeled & MRE (mm) $\downarrow$ \\ 
\midrule	
SCN~\cite{PSB19} & UNet & - & 0.66\\
GU2Net~\cite{ZYX21} & UNet & - & 0.84\\
HM$^{\dagger}$~\cite{XWW18} & ResNet-50 & - & 0.68\\
TF$^{\dagger}$~\cite{LJL22} & ResNet-50 & - & 0.69\\
\midrule	
Ours w/ HM & ResNet-50 & 12K & \underline{0.64} \\
Ours w/ TF & ResNet-50 & 12K & \textbf{0.61} \\
\bottomrule
\end{tabularx}
\label{tab:results_omni_hand}
\end{table}

\begin{figure}[t]
\centering
  \includegraphics[width=1\linewidth]{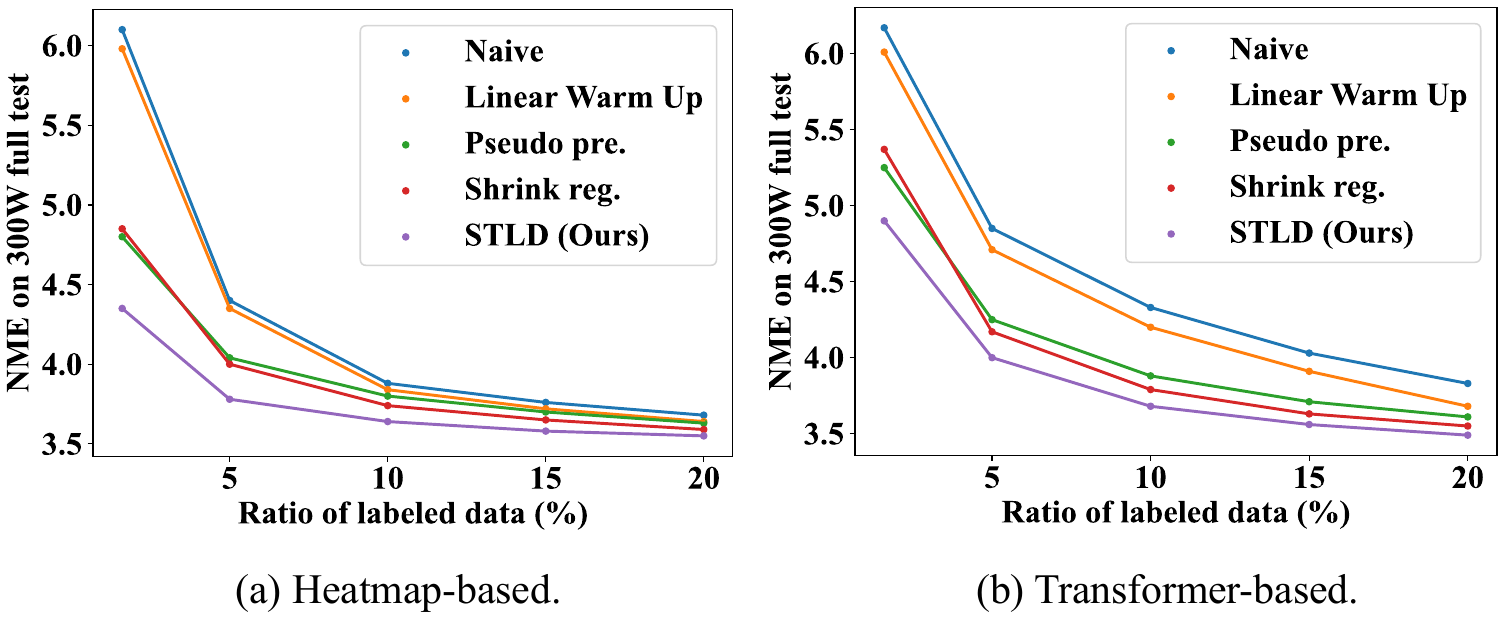}
\caption{Ablation study on 300W with different labeled ratios for (a) heatmap-based and (b) transformer-based models. The naive baseline simply utilizes all the pseudo-labels and the linear warm up baseline linearly increases the pseudo-labeled loss from 0.1 to 1 over self-training rounds.} 
\label{fig:ablation}      
\end{figure}

\subsubsection{Omni-supervised Setting}
\label{sec5.3.2}

\textbf{Facial landmark.}   
We further compare STLD with the existing methods on 300W in the omni setting, where all the methods are trained on the fully labeled training data of 300W and extra unlabeled images. To compare with previous methods that use different sizes of unlabeled images, we report results with 20K, 70K, and 200K unlabeled from CelebA~\cite{LLW15}. From Tab.~\ref{tab:results_omni_300w} that shows the relevant results, we can see that STLD-TF achieves 3.14 of NME (\%) with 200K unlabeled, which is the best among all the methods. Suprisingly, STLD-TF with only 20K unlabeled images (3.20) has surpassed several methods that use 200K, e.g., CL (3.26) and STC (3.23).

\noindent \textbf{Medical landmark.}
Compared to facial landmark, medical landmark detection benefits more from omni learning as its labeled data is usually limited due to expensive labeling cost. Therefore, we compare the omni learning results with the existing supervised methods~\cite{PSB19,ZYX21} in Tab.~\ref{tab:results_omni_hand} to see the improvement from using extra unlabeled images. As can be seen, without extra unlabeled data, HM and TF obtain 0.68 and 0.69 of MRE (mm), respectively, both of which are slightly worse than SCN~\cite{PSB19} (0.66). By leveraging extra 12K unlabeled images from RSNA~\cite{HPC19}, the results of HM and TF are boosted to 0.64 and 0.61, respectively, where the latter outperforms SCN by a large margin (i.e., 7.5\% improvement). The above results show that it is promising to boost medical landmark detection models through omni-supervised learning to deal with insufficient labeled data.  

\subsection{Model Analysis}
\label{sec5.4}

\noindent \textbf{Ablation study.}
To see the effectiveness of pseudo pretraining and shrink regression separately, we do ablation study on 300W for both HM and TF, and the results are shown in Fig.~\ref{fig:ablation}a and~\ref{fig:ablation}b, respectively. In addition to the naive baseline that simply uses all the pseudo-labels, we add another baseline named linear warm up, which linearly increases the weight of the pseudo-labeled loss from 0.1 to 1 over self-training rounds. First of all, we can see that linear warm up only outperforms the naive baseline marginally, especially with low labeled ratios, implying that the warm up strategy could not leverage pseudo-labels in an effective way. On the other hand, the proposed components both improve over the naive and linear warm up baseline consistently. The difference is that pseudo pretraining works better with small data regime, while shrink regression is better with higher labeled ratios. This is reasonable because pseudo pretraining does not directly fit the pseudo-labels, thus being more noise-resistant; in contrast, shrink regression is designed to directly leverage the pseudo-labels for better results. By combining the two, STLD obtains the best performance across all the ratios of labeled data for both models. 

\begin{figure}[t]
\centering
  \includegraphics[width=1\linewidth]{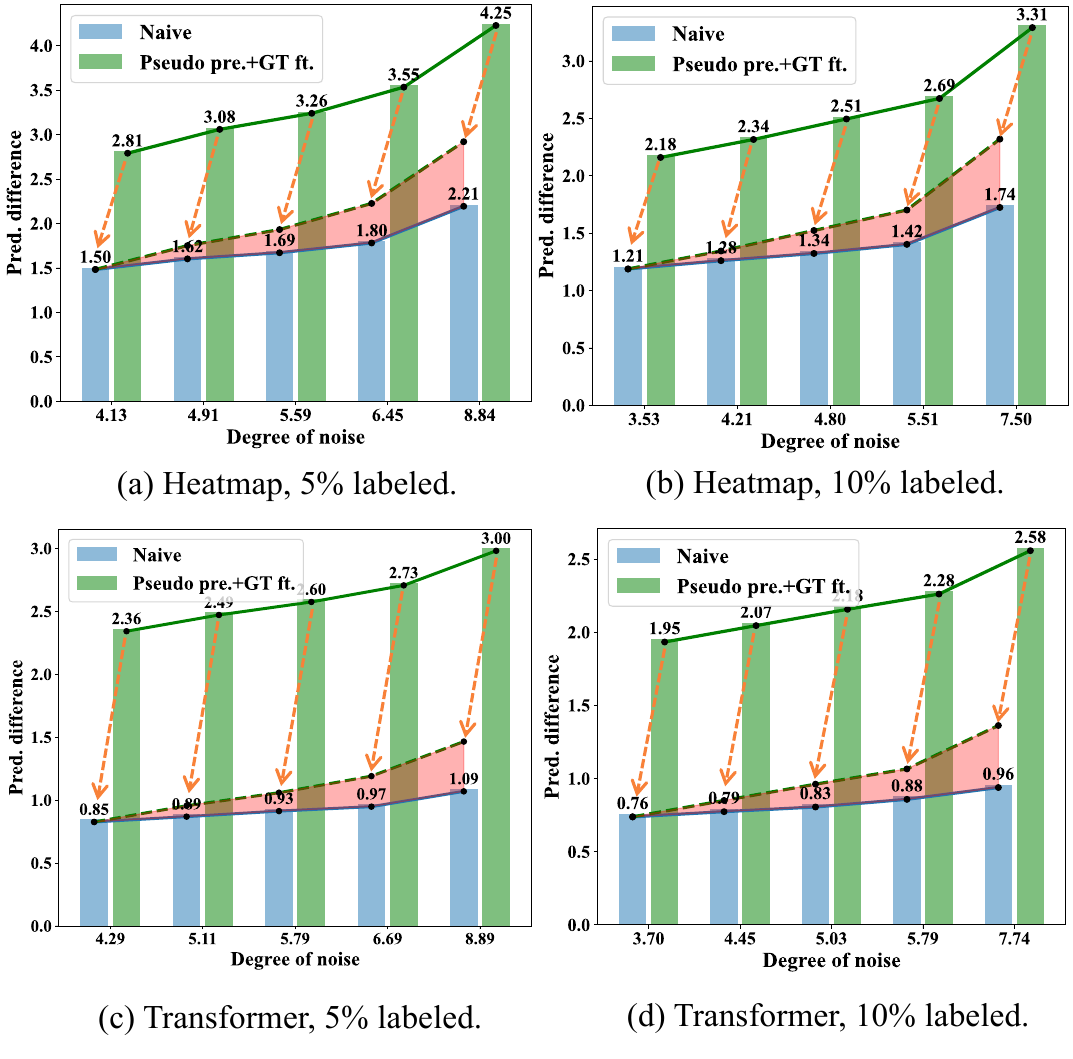}
\caption{Analysis of example forgetting by comparing the naive method and pseudo pretraining on prediction difference of pseudo-labels. (top) Heatmap-based model with 5\% and 10\% labeled data, respectively. (bottom) Transformer-based model with 5\% and 10\% labeled data, respectively.} 
\label{fig:pp_delta}      
\end{figure}

\begin{table*}[t]
\centering
\newcolumntype{C}{>{\centering\arraybackslash}X}%
\caption{Comparison between MAE and pseudo-pretraining (PP) on 300W.}
\begin{tabularx}{\linewidth}{lCCCC}
\toprule	 
Method & 1.6\% & 5\%  & 10\% & 20\%  \\ 
\midrule	
No Pretrain & 4.26 & 3.64 & 3.48 & 3.34 \\
MAE~\cite{HCX22} & 4.19(-1.6\%) & 3.60(-1.1\%) & 3.45(-0.9\%) & 3.33(-0.3\%) \\
PP (Ours) & 3.97(-\textbf{6.8\%}) & 3.47(-\textbf{4.7\%}) & 3.39(-\textbf{2.6\%}) & 3.30(-\textbf{1.2\%}) \\
\bottomrule
\end{tabularx}
\label{tab:mae_pseudo}
\end{table*}

\begin{figure}[t]
\centering
  \includegraphics[width=1\linewidth]{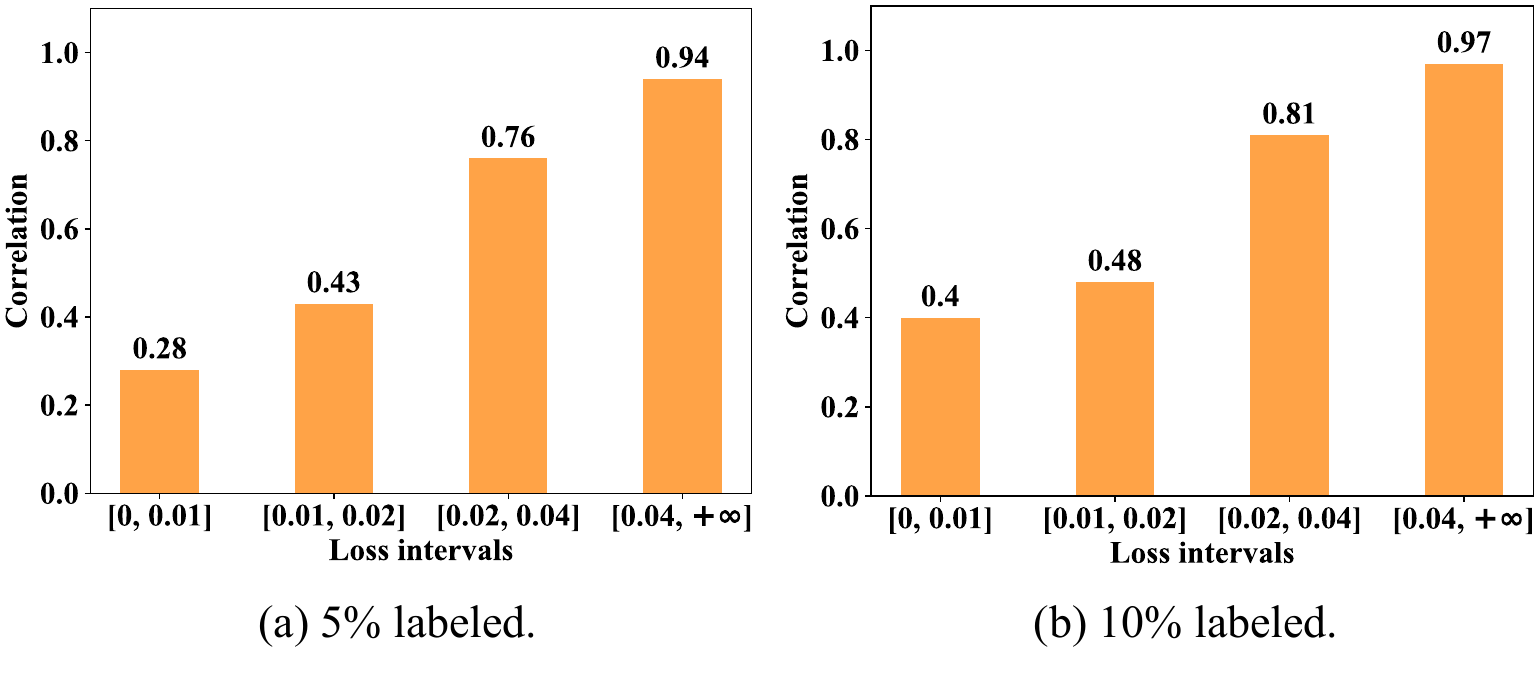}
\caption{Pearson correlation coefficient of the gradients between pseudo-labels and GTs of TF model, trained on 300W. (a) 5\% labeled. (b) 10\% labeled.} 
\label{fig:corr}      
\end{figure}

\begin{figure}[!t]
\centering
  \includegraphics[width=0.85\linewidth]{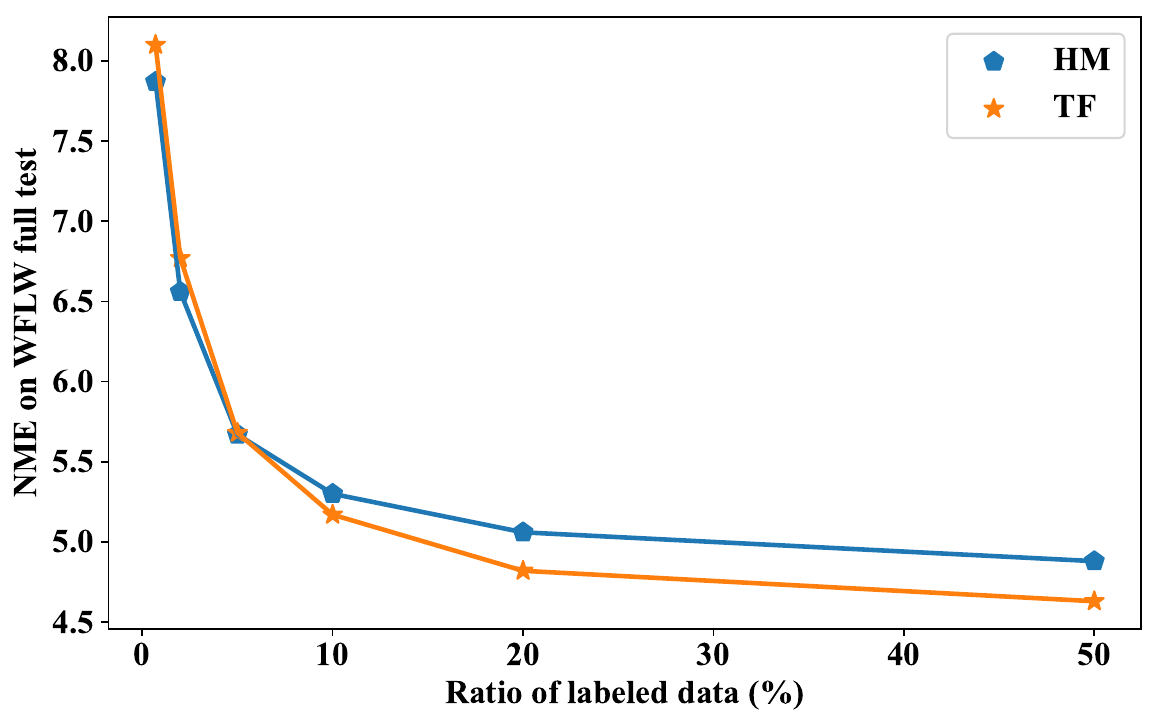}
\caption{NME performance of HM and TF models on WFLW with various labeled ratios. } 
\label{fig:inductive_bias}      
\end{figure}

\noindent \textbf{How pseudo pretraining works.}
One may worry that the noisy samples learned during pseudo pretraining would be memorized by the model and thus leads to confirmation bias. To understand how it works, we perform an analysis of example forgetting by computing the difference of the pseudo-labels used for training and their predictions after the two-stage learning. Note that shrink regression is not used in the second stage here as we would like to focus on pseudo pretraining. As a comparison, we also do the analysis for the naive baseline. Fig.~\ref{fig:pp_delta} shows the relevant results of both HM (top) and TF (bottom), with different labeled ratios, and the pseudo-labels are grouped by noise. First of all, we can see that the naive baseline forgets more about noisier samples, which is consistent to the findings from~\cite{TSC19}. And as expected, the model with the two-stage training does not memorize the pseudo-labels as much as the naive baseline does because the pseudo-labels are not used in the second stage of the proposed method. But interestingly, the forgetting of noisier samples is enlarged when compared to the naive baseline (i.e., the pink area\footnote{The green dotted line of the pink area is obtained by moving the upper green solid line down and overlaps with the lower green solid line at the first black dot.}), indicating that the two-stage design with pseudo pretraining is able to handle noisy pseudo-labels implicitly. Such a characteristic helps pseudo pretraining to effectively utilize the task-specific knowledge of pseudo-labels while avoiding the influence of noisy samples. To show the superiority of pseudo pretraining, we compare it with masked autoencoder (MAE)~\cite{HCX22}, a popular self-supervised pretraining method. For reference, we add a baseline named No Pretrain, which is directly fine-tuned on labeled data. Tab.~\ref{tab:mae_pseudo} gives the results. We can see that pseudo-pretraining outperforms MAE consistently across all the labeled ratios, verifying the advantage of utilizing pseudo-labels over unlabeled pretraining as the former incorporates task-specific knowledge for better performance. 

\begin{figure}[!t]
\centering
  \includegraphics[width=1\linewidth]{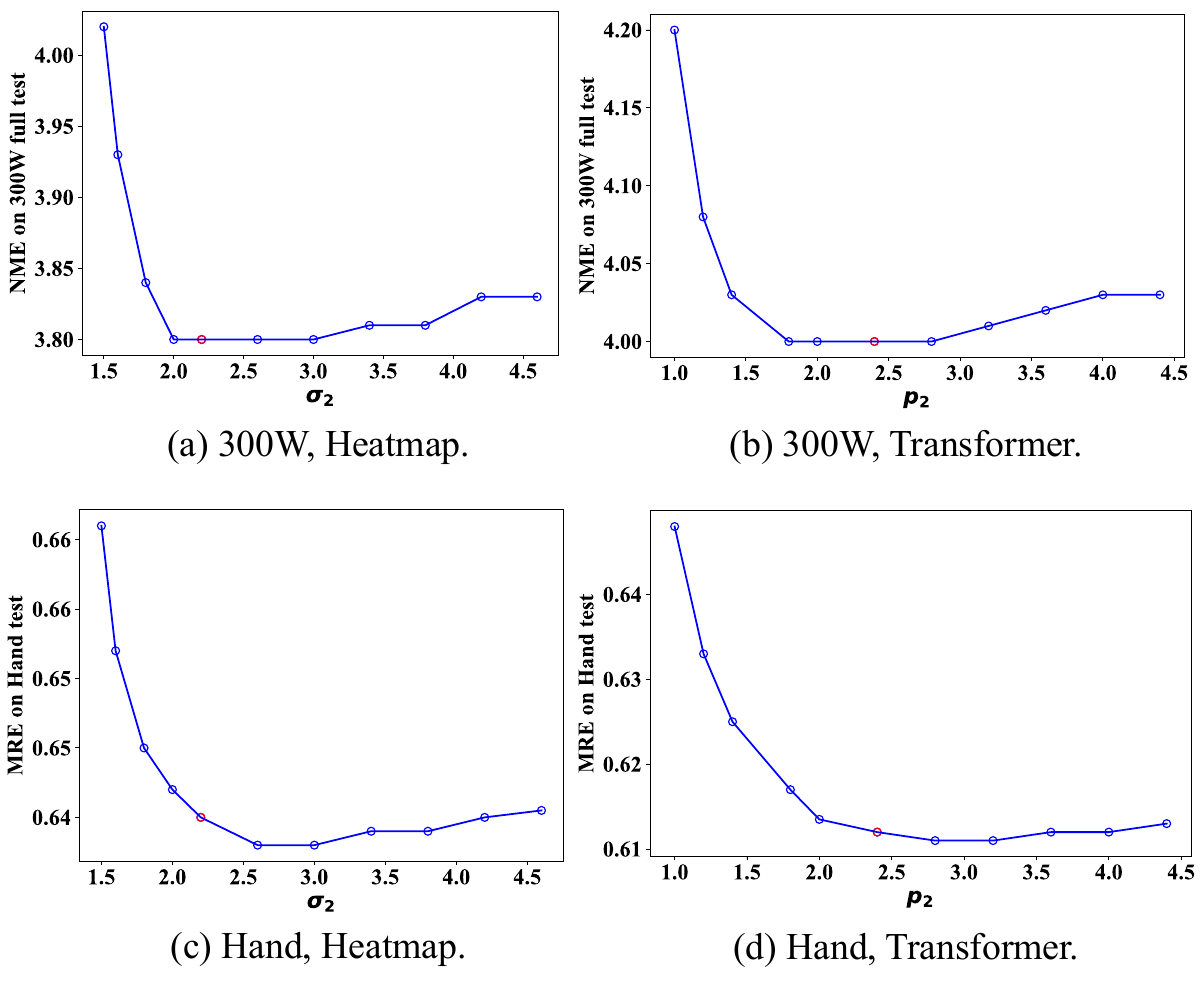}
\caption{Model performance with different curriculum settings of shrink regression. \textcolor{red}{Red dot} indicates the setting used in the experiments. (a)-(b): NME (\%) on 300W test of heatmap and transformer model, respectively, under semi-supervised learning with 5\% labeled. (c)-(d): MRE (mm) on Hand test of heatmap and transformer model, respectively, under omni-supervised learning.} 
\label{fig:curri_range}      
\end{figure}

\noindent \textbf{How shrink loss works.}
While the shrink regression for heatmap-based models looks straightforward, the shrink loss applied to coordinate-based models seems not so obvious on adjusting task confidence. To verify it, we perform a correlation analysis of the gradients between GTs and pseudo-labels. We record the gradients\footnote{The seeds are fixed to make sure each pair of gradients from GTs and pseudo-labels is generated by the same batch of training data.} of the last layer when trained with GTs and pseudo-labels, respectively, and group them by pseudo-label noise. For each group, we compute the Pearson correlation of the gradients between GTs and pseudo-labels. Fig.~\ref{fig:corr}a and~\ref{fig:corr}b show the results on 5\% and 10\% labeled settings of 300W, respectively. We can see that the correlation decreases as the loss scale decreases, which indicates that the gradients of small errors trained on pseudo-labels are indeed noisier and less reliable. Therefore, it is reasonable to increase the confidence of the regression task by downweighting the gradients of small prediction errors.  

\begin{figure}[t]
\centering
  \includegraphics[width=0.96\linewidth]{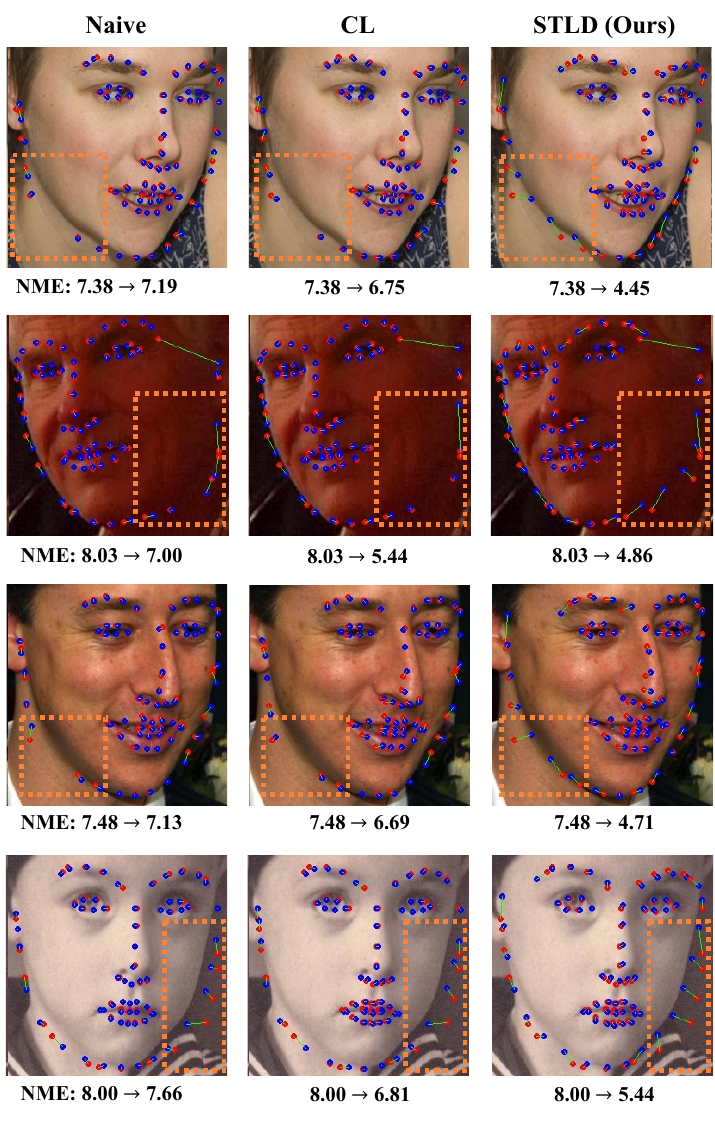}
\caption{Visualized unlabeled samples from 300W, with pseudo-labels from the first (\textcolor{red}{red dots}) and the last self-training round (\textcolor{blue}{blue dots}). The naive method (left), CL (middle), and STLD (right) all use the same base model HM, trained with 1.6\% labeled data. The NME is computed between the GTs and pseudo-labels.} 
\label{fig:vis_compare}      
\end{figure}

\begin{table*}[t]
\centering
\small
\newcolumntype{C}{>{\centering\arraybackslash}X}%
\caption{Comparison between self-training and our method in computational cost. The training time represents the total training time of a model while the inference time represents the time cost for each test image. One RTX 3090 GPU was used in both settings.}
\begin{tabularx}{\linewidth}{lCCCCCC}
\toprule	 
 & \multicolumn{3}{c}{Training} & \multicolumn{3}{c}{Inference} \\ \cmidrule(r){2-4} \cmidrule(r){5-7}
Backbone & Self-training & STLD-HM & STLD-TF & Self-training & STLD-HM & STLD-TF \\
\midrule
ResNet-18 & 0.75 hour & 1.00 hour & 6.00 hours & 3.25 ms & 3.34 ms & 2.61 ms \\
HRNet & 3.20 hours & 3.50 hours & 22.00 hours & 4.34 ms & 4.35 ms & 6.39 ms \\
\bottomrule
\end{tabularx}
\label{tab:comp_cost}
\end{table*} 

\begin{figure}[t]
\centering
  \includegraphics[width=0.95\linewidth]{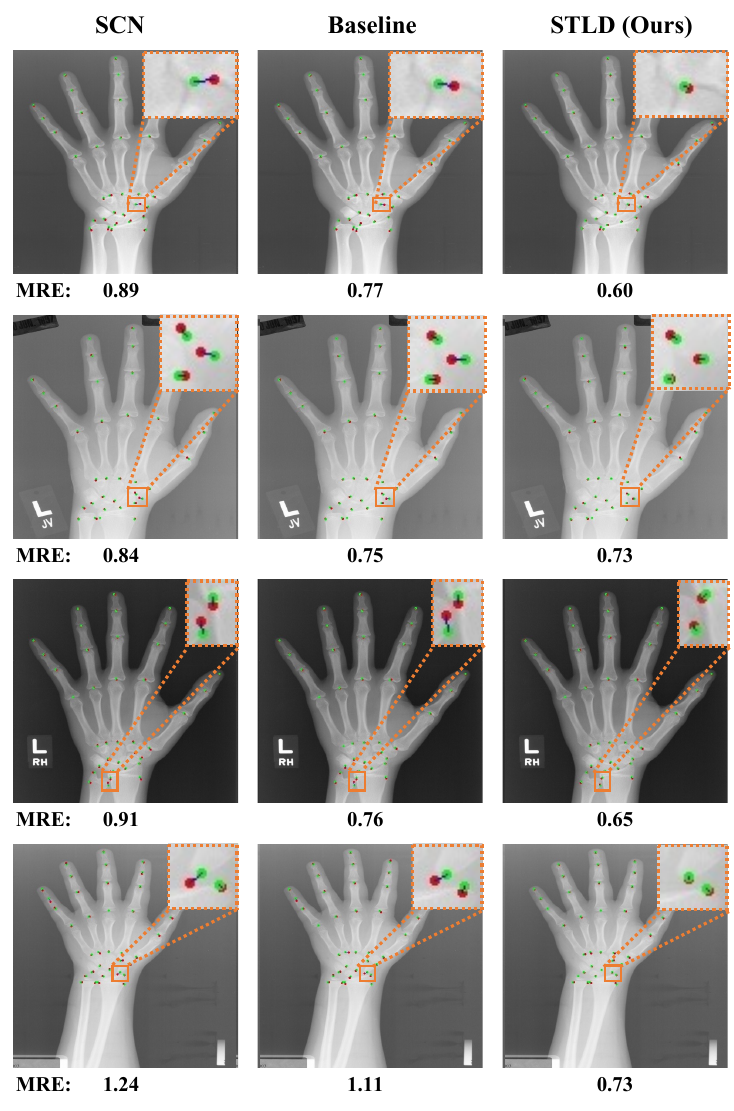}
\caption{Visualized test samples from Hand, with ground-truths (\textcolor{green}{green dots}), predictions (\textcolor{red}{red dots}), and the distances between them (\textcolor{blue}{blue lines}). SCN (left) and the baseline HM (middle) are trained in supervised learning while STLD (right) is trained in omni-supervised learning.} 
\label{fig:vis_hand}      
\end{figure}

\noindent \textbf{HM vs. TF.}
The two variants of STLD, HM and TF, both deliver promising results on popular benchmarks (see Tab.~\ref{tab:results_semi}). However, we observe different characteristics of the two in data efficiency and performance ceiling. Specifically, HM obtains better performance with smaller labeled ratios while TF performs better when the size of labeled data increases. This observation is consistent to that of prior works~\cite{CLJ19,DBK20,DTL21}, where CNNs are known to be data-efficient due to stronger inductive bias while transformers have a higher performance ceiling. Since prior works are mostly based on classification, we compare the data efficiency and performance trade-off of HM and TF in landmark detection under semi-supervised learning, which is given in Fig.~\ref{fig:inductive_bias}. The figure plots the NME performance of the two models with various labeled ratios on WFLW. We can see that the turning point is at around 5\% labeled ratio (i.e., 375 samples), where TF starts to show better performance than that of HM model due to its weaker assumption of inductive bias. This gives us a guideline on the selection of STLD variant in practice, where the HM model is preferred if the available labeled data is quite limited. Alternatively, one may consider incorporating inductive bias into transformers~\cite{DTL21} or applying distillation strategies~\cite{TCD21} to enhance the performance of TF model in the small-data regime, which is beyond the scope of the paper.

\noindent \textbf{Curriculum sensitivity.}
To investigate the sensitiveness of the curriculum of shrink regression, we conducted experiments with different granularity curriculum settings (i.e., the values of [$\sigma_2$, $\sigma_3$, $\sigma_4$] and [$p_2$, $p_3$, $p_4$] for HM and TF, respectively). Since the granularity becomes the same as standard regression in the last round (i.e., $\sigma_4=1.5$ and $p_4=1$), we vary the curriculum by adjusting $\sigma_2$/$p_2$ while setting $\sigma_3$/$p_3$ to approximately the middle value of $\sigma_2$/$p_2$ and $\sigma_4$/$p_4$. Fig.~\ref{fig:curri_range}a and~\ref{fig:curri_range}b give the NME results on 300W with different $\sigma_2$/$p_2$ of HM and TF, respectively. We can see from the figures that the performance remains largely stable across a wide range of $\sigma_2$/$p_2$ for both models. For example, HM obtains the optimal performance of 3.81 in NME (\%) when $\sigma_2$ ranges from 2.0 to 3.0, and the performance slightly drops in the range [3.0, 4.6]. When $\sigma_2$ approaches 1.5 (i.e., the value used for standard regression), the performance drops significantly because the coarse-to-fine curriculum of shrink regression disappears, leading to confirmation bias. The situation is similar for the TF model. Therefore, shrink regression is relatively robust to different settings of the granularity curriculum. To see whether the selected curriculum based on 300W can be directly applied to a quite different dataset, we analyze its sensitivity on the Hand dataset, which is shown in Fig.~\ref{fig:curri_range}c and~\ref{fig:curri_range}d. We can see that the selected $\sigma$/$p$ using 300W also obtains strong performance on Hand, indicating the robustness of the curriculum. Undeniably, the optimal curriculum of Hand is slightly larger than that of 300W, which might be due to the larger noise in the pseudo-labels of Hand. Overall, the tuned curriculum performs robustly across datasets and does not need to be tuned for each dataset.

\noindent \textbf{Qualitative results.}
Lastly, we show the superiority of STLD via qualitative results. We first show its advantage in pseudo-label refinement in Fig.~\ref{fig:vis_compare} using 300W. The results of the naive baseline (left) and CL (middle) are also given for comparison. For each image, we plot the initial (\textcolor{red}{red dots}) and the last round pseudo-labels (\textcolor{blue}{blue dots}) to see the improvement. We can see that CL performs better than the naive method in general, and STLD is the best among the three. To be specific, in the first row, the naive method and CL barely change the pseudo-labels inside the rectangle area due to confirmation bias, while STLD is able to largely refine them and thus reduces the NME from 7.38 to 4.45. In the second row, both the naive method and CL are able to refine the pseudo-labels, but the quality is unsatisfactory; on the other hand, our method obtains high-quality pseudo-labels in the last round, as evidenced by the visualized predictions and the significant improvement of NME (from 8.03 to 4.86). To summarize, the baseline and CL are not able to obtain high-quality pseudo-labels due to either confirmation bias (row 1 and 3) or unsatisfactory estimates (row 2 and 4). In contrast, the proposed STLD handles confirmation bias properly through the task curriculum, which further boosts the quality of the predicted pseudo-labels. We then show the advantage of STLD in accurate prediction in Fig.~\ref{fig:vis_hand} using Hand test samples. The SCN and HM baseline are included as reference. We plot ground-truths (\textcolor{green}{green dots}), model predictions (\textcolor{red}{red dots}), as well as the distance (\textcolor{blue}{blue lines}) between them for easy comparison. We can see from the figures that STLD makes more accurate predictions than the two counterparts consistently, which validates the promise of applying STLD with omni-supervised learning for boosting medical landmark detection.

\noindent \textbf{Computational cost.} 
Since the proposed method requires more training stages than a standard self-training method, the computational cost can be expensive. To investigate this, we compare the training time as well as inference time of our method with self-training~\cite{Lee13}. We evaluate the total training time on 300W with 10\% labeled data; for inference, we evaluate the inference time per image. Both training and inference were conducted with one RTX 3090 GPU. Table~\ref{tab:comp_cost} shows the computational cost of the two methods in different backbones for both training and inference. It can be seen that our STLD-HM model has a slightly longer training time than that of self-training for both ResNet-18 and HRNet, indicating that the training complexity of our method is acceptable as the two use the same base detector (i.e., heatmap regression). However, STLD-TF requires much longer training time (about 7 times), due to the slow convergence of transformers~\cite{CMS20}. Such an issue might be mitigated by faster adaptations such as deformable detr~\cite{ZSL21}, which spends 10 times less training epochs than the standard framework~\cite{CMS20}. For inference speed, STLD-HM and self-training are again comparable. Surprisingly, STLD-TF with ResNet-18 is faster than the heatmap counterparts and we believe this is because coordinate regression does not require upsampling, which saves time. On the other hand, STLD-TF with HRNet is slower than the other two and we attribute it to the high-resolution feature map of HRNet, which significantly increases the computation of the self-attention in transformers.

\section{Conclusion and Future Work}
\label{sec6} 

In this work, we proposed STLD, a two-stage self-training method for semi-supervised landmark detection. Without explicit pseudo-label selection, the method is still able to leverage pseudo-labels effectively by constructing a task curriculum that transitions from more confident to less confident tasks. Compared to selection-based self-training, the advantages of STLD are three-fold: 1) it does not introduce data bias caused by sample selection; 2) it avoids the choice of confidence threshold that can be sensitive; 3) it is applicable to coordinate regression methods that do not output confidence scores. Experiments on four landmark detection benchmarks have demonstrated the effectiveness of the method in both semi- and omni-supervised settings. We highlight that the contribution of the paper is not limited to the proposal of STLD, but also a new way of applying self-training. It would be interesting to see more instantiations of the task curriculum that works for landmark detection and beyond (e.g., object detection and semantic segmentation), which we leave for future work.

\bibliographystyle{IEEEtran}
\bibliography{mybib}

\vfill

\end{document}